%% file: main.tex
\documentclass[10pt,twocolumn,letterpaper]{article}

\usepackage[pagenumbers]{wacv} 

\input{preamble}

\definecolor{wacvblue}{rgb}{0.21,0.49,0.74}
\usepackage[pagebackref,breaklinks,colorlinks,allcolors=wacvblue]{hyperref}

\def\wacvPaperID{1206} 
\def\confName{WACV}
\def\confYear{2027}

\title{SWAP-INR: Implicit Neural Representation with Switchable Attention Paths \\ for Time-varying Multivariate Scientific Data Compression}

\author{
Hyunsoo Son\textsuperscript{1$\ast$} \quad
Jeonghyun Noh\textsuperscript{1$\ast$} \quad
Chaoli Wang\textsuperscript{2} \quad
Won-Ki Jeong\textsuperscript{1$\dagger$}\\[0.4em]
\textsuperscript{1}Korea University \qquad
\textsuperscript{2}University of Notre Dame\\[0.3em]
{\tt\small
\textsuperscript{1}\{ehsson,wjdgus0967,wkjeong\}@korea.ac.kr \quad
\textsuperscript{2}chaoli.wang@nd.edu
}
}

\begin{document}

\makeatletter
\let\old@maketitle\@maketitle
\renewcommand{\@maketitle}{%
  \old@maketitle%
  \begin{center}
    \includegraphics[width=0.94\linewidth,keepaspectratio]{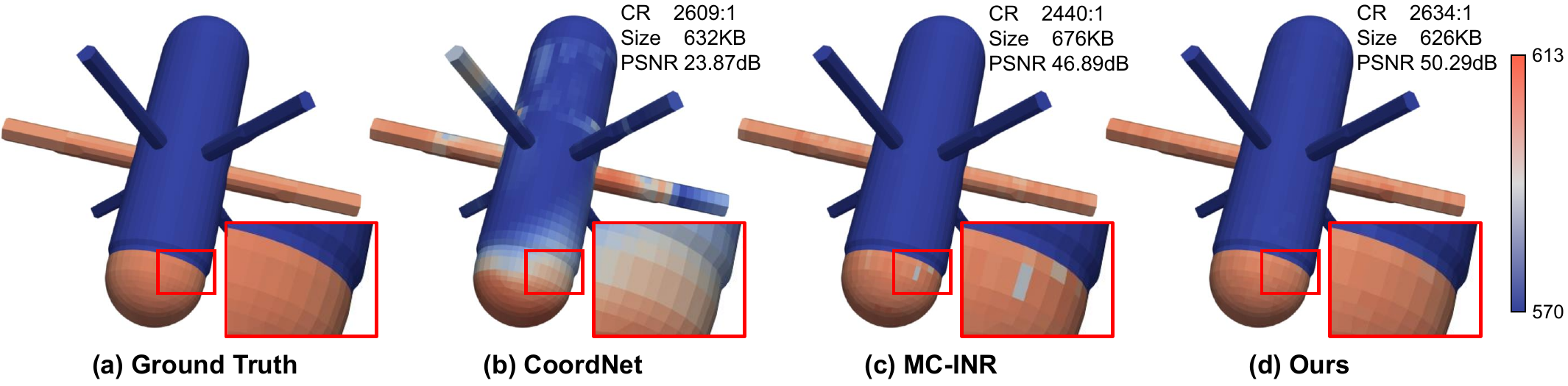}
    \captionof{figure}{
    Visual comparison of compression quality on the CUPID~\cite{cho2019numerical} at matched compression ratios.
    (a) Ground truth.
    (b) CoordNet~\cite{han2022coordnet} struggles to capture multivariate characteristics, resulting in a significant loss of structural details.
    (c) MC-INR~\cite{son2025mc} isolates variables, missing explicit cross-variable interactions, and produces severe visual artifacts (red box).
    %
    (d) Our proposed \textbf{SWAP-INR} preserves high-fidelity physical details with superior PSNR by switching between the attention and non-attention paths to capture essential dynamics.
    }
    \label{fig:teaser}
  \end{center}
}
\makeatother

\maketitle

\begingroup
\renewcommand{\thefootnote}{$\ast$}
\footnotetext{ denotes equal contribution, and $^\dagger$ means the corresponding author.}
\endgroup

\input{_sec/0_abstract}    
\input{_sec/1_introv2}
\input{_sec/2_relatedv2}
\input{_sec/3_method}
\input{_sec/4_exp}
\input{_sec/5_con}

\section*{Acknowledgments}
This work was supported in part by the National Research Foundation of Korea under Grant RS-2024-00349697 and Grant RS-2021- NR060143; in part by the Institute for Information and Communications Technology Planning and Evaluation under Grant IITP-2027-RS-2020-II201819; in part by the National Research Council of Science \& Technology(NST) grant by the Korea government (MSIT) (No. GTL24033-000); and in part by Korea University Grant. C. Wang is supported by the U.S. National Science Foundation through grants IIS-2101696, OAC-2104158, IIS-2401144, and CCF-2550610.

{
    \small
    \bibliographystyle{ieeenat_fullname}
    \bibliography{main}
}

\newpage
\appendix
\input{_sec/6_sup}

\end{document}

%% file: preamble.tex
\newcommand{\figref}[1]{Fig.~\ref{#1}}

\newcommand{\tabref}[1]{Tab.~\ref{#1}}

\newcommand{\Figref}[1]{Figure~\ref{#1}}

\newcommand{\Tabref}[1]{Table~\ref{#1}}

\definecolor{watermelon_green}{rgb}{0, 0.42, 0.30}

\usepackage{graphicx,verbatim}

\usepackage{verbatim}
\usepackage{graphicx}
\usepackage{bibentry}
\usepackage{amsmath}
\usepackage{multirow}
\usepackage{tabularx}
\usepackage{makecell}
\usepackage{bbding}
\usepackage{pifont}
\usepackage{wasysym}
\usepackage{amssymb}
\usepackage{svg}
\usepackage[table]{xcolor}
\usepackage{rotating}
\usepackage[export]{adjustbox}
\usepackage{anyfontsize}

%% file: _sec/0_abstract.tex
\begin{abstract}
Visualizing multivariate scientific simulation data is essential for understanding complex physical phenomena, yet the massive scale of such data imposes a severe storage bottleneck in high-performance computing environments.
Implicit neural representations (INRs) have emerged as a promising continuous alternative to traditional lossy compressors, but existing methods struggle to explicitly capture cross-variable interactions and to adapt to the diverse spectral characteristics of physical variables.
To address these limitations, we propose \textbf{SWAP-INR}, an architecture for high-fidelity compression of multivariate, time-varying simulation data. 
SWAP-INR introduces two key components:
(1) The switchable attention path (SWAP) module explicitly models cross-variable interactions through a cross-variable attention layer, then routes each variable to an attention or non-attention path by a deterministic, loss-driven criterion, and prunes the inactive layers for structural compression.
(2) The adaptability amplification SIREN (A$^2$-SIREN) replaces rigid frequency scaling with channel-wise learnable modulation, adapting to heterogeneous physical features with negligible overhead.
Combined with post-training 16-bit quantization, 
%
SWAP-INR achieves state-of-the-art performance on small modular reactor (SMR) datasets and remains competitive on public physics simulation datasets, while maintaining a constrained memory footprint.
%
Our code is \url{https://github.com/hvcl/SWAP-INR}.
\end{abstract}

%% file: _sec/1_introv2.tex
\section{Introduction}
\label{sec:intro}

Scientific simulation is central to studying complex real-world phenomena that are too costly, dangerous, or inaccessible to observe directly, and visualizing the resulting data is essential for analysis and decision-making~\cite{shen2006visualization,muigg2007scalable,yu2010situ,wang2018visualization}.
In safety-critical engineering domains such as small modular reactor (SMR) design, high-fidelity simulations couple multiple physical processes, including fluid dynamics, heat transfer, and thermodynamics.
The resulting data are multivariate and time-varying, defined at high spatial and temporal resolution on unstructured grids that conform to complex geometries.
Such fields encode rich physical behavior and thus accumulate substantial fine-grained detail, so storing or transmitting them at full resolution exceeds the memory and bandwidth budgets of high-performance computing (HPC) environments.
Overcoming this bottleneck requires representations that substantially reduce memory consumption while preserving the fidelity needed for accurate analysis.

To alleviate this storage bottleneck, hand-crafted lossy compressors (TTHRESH~\cite{ballester2019tthresh}, SZ3~\cite{liang2022sz3}, and ZFP~\cite{lindstrom2014fixed}) have been widely adopted for scientific simulation data.
However, their reliance on fixed transforms, predictors, and quantization schemes limits their adaptability to complex, dataset-specific structures.
Recently, implicit neural representations (INRs) have emerged as a promising alternative, encoding scientific data as continuous functions and supporting queries at arbitrary spatial locations.
Following the SIREN framework~\cite{sitzmann2020implicit}, coordinate-based methods such as CoordNet~\cite{han2022coordnet} and its variants~\cite{tang2023ecnr,han2023kd,son2025mc} have 
demonstrated strong compression performance while supporting continuous queries at arbitrary spatial locations.
In parallel, grid-based methods, such as fv-SRN~\cite{weiss2022fast}, APMGSRN~\cite{wurster2023adaptively}, and Wu~\etal~\cite{wu2024interactive}, employ multi-scale feature grids to encode spatially located data more effectively.

Despite their effectiveness, existing INR-based methods exhibit several limitations when applied to multivariate scientific simulation data.
%
First, most INRs are designed for single-variable fields and handle multiple variables by simply expanding the output channels of a shared network.
%
Since the representation is fully shared, distinct variables compete for the same parameters and impose conflicting gradients, making it difficult for the model to learn inter-variable relationships.
%
This may cause feature interference that degrades decoding quality, a gradient-conflict effect studied in shared-parameter optimization~\cite{yu2020pcgrad,standley2020which}.
%
The recent MC-INR~\cite{son2025mc} partially mitigates this limitation with a branched architecture, but it merely decouples variables into separate branches without explicitly modeling cross-variable dependencies.
%
Moreover, because the relevant interactions among variables are unknown a priori and dataset-dependent, a fixed architecture is inherently limited in its ability to capture these dependencies.
Second, existing INRs often struggle to adapt efficiently to the diverse physical characteristics present within a single dataset. 
%
Baselines such as SIREN~\cite{sitzmann2020implicit} are efficient, but their fixed frequency scaling limits the representation of variables with heterogeneous spectral properties.
Remedies that increase network capacity~\cite{lindell2022bacon,zhang2023frequency} or partition the spatial domain~\cite{saragadam2022miner} improve adaptability at substantial computational overhead.

To address these limitations, we propose \textbf{SWAP-INR}, an INR network tailored for high-fidelity compression of multivariate, time-varying scientific data.
Extending coordinate-based INRs, our method introduces two core components.
First, the switchable attention path (\textbf{SWAP}) module models multivariate interactions through a cross-variable attention layer, \emph{automatically discovers which variables should interact}, and then dynamically routes each variable through an attention or non-attention path rather than relying on a hand-specified correlation structure.
This routing prevents forced correlations among non-interacting variables, and the unselected layers are then pruned to yield structural compression.
Second, adaptability amplification SIREN (\textbf{A$^2$-SIREN}) replaces rigid frequency scaling with neuron-level learnable modulation, resolving heterogeneous physical features at negligible computational cost.
Combined with post-training 16-bit quantization, SWAP-INR delivers high-fidelity representations within a constrained memory budget.
%
As a result, SWAP-INR outperforms existing methods on small modular reactor (SMR) datasets and achieves comparable results on public simulation datasets, at matched compression ratios and PSNR settings, preserving the underlying structures (\figref{fig:teaser}).
Our main contributions are summarized as follows:

\begin{itemize}
    \item We propose the SWAP module, which explicitly models cross-variable interactions via a cross-variable attention layer.
    %
    By dynamically switching paths and pruning inactive layers, it eliminates feature interference and achieves structural compression.
    %

    \item We introduce A$^2$-SIREN, which relaxes the rigid global frequency modulation of traditional INRs.
    Its neuron-level learnable modulation adaptively captures data-dependent high-fidelity physical details with negligible computational overhead.
    
    \item We demonstrate that the proposed SWAP-INR achieves state-of-the-art (SOTA) performance on in-house SMR datasets and competitive results on public datasets.
    
\end{itemize}

%% file: _sec/2_relatedv2.tex
\section{Related Work}
\label{sec:related}

\input{_fig/fig_tex/fig2_overview}

\subsection{Implicit Neural Representations}

INRs based on standard ReLU-multi-layer perceptrons (ReLU-MLPs) suffer from severe spectral bias, failing to capture high-frequency details~\cite{tancik2020fourier}.
%
Two paradigms address this.
The first advances activation functions.
%
SIREN~\cite{sitzmann2020implicit} introduced periodic sine activations, on which BACON~\cite{lindell2022bacon} imposed progressively expanding frequency bandwidths, and AFNet~\cite{zhang2023frequency} assigned localized frequency scaling via an auxiliary network.
%
Other works replace the activation itself with wavelet~\cite{saragadam2023wire}, variable-periodic~\cite{liu2024finer}, or polynomial~\cite{rezaeian2025sl2a} formulations, or re-parameterize training in the Fourier domain~\cite{shi2024improved}.
The second paradigm encodes spatial and spatiotemporal coordinates. Positional encoding~\cite{mildenhall2021nerf}, hierarchical input partitioning~\cite{saragadam2022miner}, and multi-resolution hash grids~\cite{muller2022instant} capture fine detail, while grid-factorization methods such as K-Planes~\cite{fridovich2023k} and HexPlane~\cite{cao2023hexplane} extend this to 4D spatiotemporal volumes.
Recent encoding schemes such as content-aware Fourier features~\cite{ke2026content} have begun to relax the dependence on fixed frequency bases.
Yet across these paradigms, frequency adaptation typically requires dedicated encoding modules, rigid grid structures, or heavy architectural changes.
In contrast, we introduce a simple yet efficient learnable channel-wise modulation that dynamically adapts to the data with negligible overhead.

\subsection{Scientific Data Compression}
\noindent\textbf{Single-variable.}
Traditional compressors operate directly on discrete data through transform-, prediction-, or decomposition-based schemes (ZFP~\cite{lindstrom2014fixed}, SZ3~\cite{liang2022sz3}, TTHRESH~\cite{ballester2019tthresh}), and thus face inherent memory limitations, with grid-based variants further constrained to structured topologies.
As a parameter-efficient alternative, NeurComp~\cite{lu2021compressive} and CoordNet~\cite{han2022coordnet} introduced INRs that map continuous spatiotemporal coordinates directly, removing the dependence on discrete grids.
Building on this, later methods improved representational capacity through hierarchical Laplacian pyramids~\cite{tang2023ecnr}, knowledge distillation~\cite{han2023kd}, spatially adaptive feature encoding~\cite{wurster2023adaptively}, and mixture-of-experts partitioning~\cite{han2026moe}.
Others focused on practical deployment: ray-tracing integration~\cite{weiss2022fast}, interactive hash encoding~\cite{wu2024interactive}, distributed \textit{in-situ} processing~\cite{wu2024distributed}, and meta-learned adaptation~\cite{yang2025meta}.
Despite this progress, all of these methods are designed for single-variable fields.

\noindent\textbf{Multivariate.}
Naively applying single-variable INRs to multivariate data forces strict weight sharing across distinct variables, causing severe feature interference~\cite{yu2020pcgrad,standley2020which}.
STSR-INR~\cite{tang2024stsr} mitigates this by conditioning a shared network on variable-specific latent vectors, but still entangles distinct dynamics.
MC-INR~\cite{son2025mc} instead encodes each variable through a separate network, isolating the variables but discarding the cross-variable correlations essential to physical systems.
To resolve this trade-off, we propose the SWAP module, which routes physical fields through switchable attention and non-attention paths, explicitly modeling cross-variable interactions while isolating independent features.

%% file: _fig/fig_tex/fig2_overview.tex
\begin{figure*}[t]
    \centering
    \includegraphics[width=0.99\textwidth]{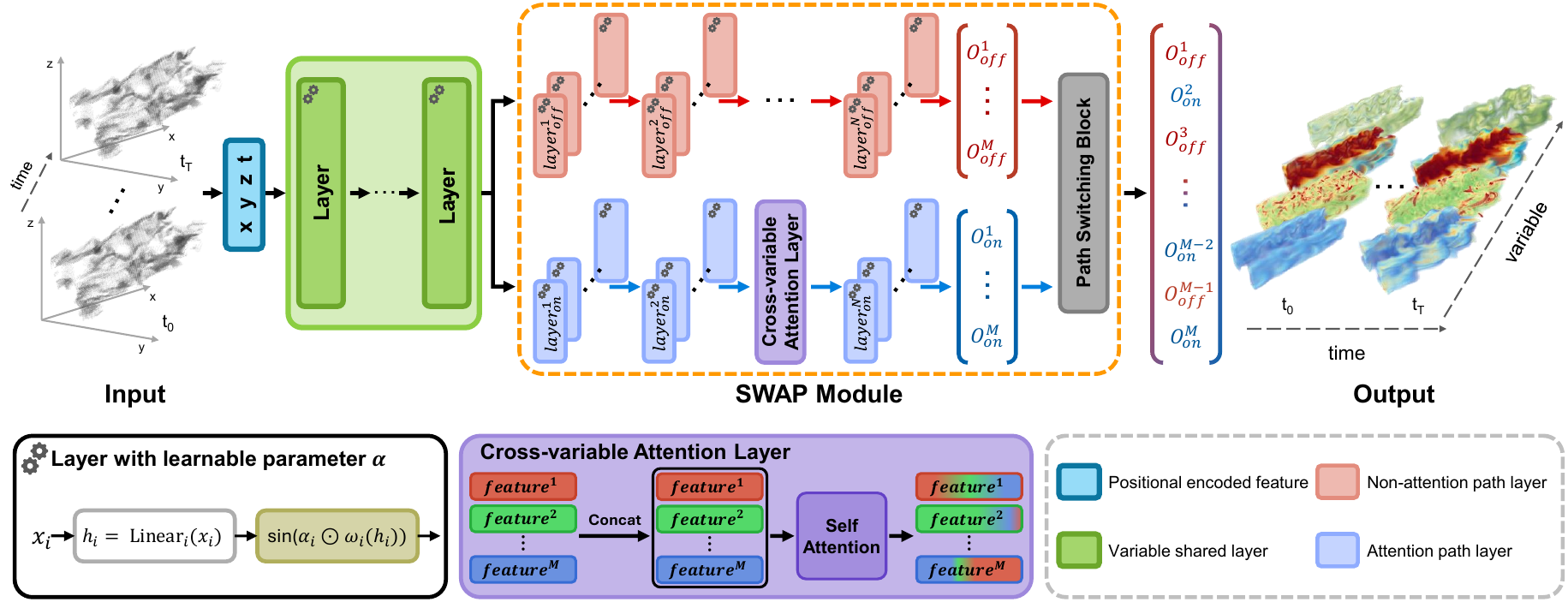}
    \caption{
    \textbf{Overview of the proposed SWAP-INR.} 
    Given a spatiotemporal coordinate $(x, y, z, t)$, the network first applies positional encoding.
    The feature is then processed through MLP layers shared across variables, equipped with A$^2$-SIREN, which uses a learnable parameter $\boldsymbol{\alpha}$ for adaptive frequency modulation (bottom left).
    The representations are split into two branches: a non-attention path (red) and an attention path (blue), where the latter models cross-variable interactions via a cross-variable attention layer (purple). 
    The variable-driven path switching block selects the optimal representation per variable to reconstruct high-fidelity multivariate outputs.
    }
    \label{fig:overview}
\end{figure*}

%% file: _sec/3_method.tex
\section{Method}

\label{sec:method}

We begin with A$^2$-SIREN, followed by the SWAP module, and finally the training scheme, loss function, and quantization procedure.
An overview of our approach is presented in \figref{fig:overview}.
As our method builds on coordinate-based INRs, it maps continuous coordinates directly to physical values, making it applicable to both structured and unstructured grids without modification.


\subsection{A$^2$-SIREN}
SIREN~\cite{sitzmann2020implicit} effectively represents the physical values using a pre-defined amplification factor ($\omega = 30$) and sine activation ($\mathtt{sin}$), formulated as:
\begin{equation}
\label{eq:siren}
    \mathbf{x}_{i+1} = \mathtt{sin}( \omega\, \mathbf{h}_i(\mathbf{x}_i)), \quad i = 0, \dots, \, N-1,
\end{equation}
%
where $\mathbf{h}_i$ is the $i$-th linear layer and $\mathbf{x}_i$ its input, with $\mathbf{x}_0$ the network input; each output $\mathbf{x}_{i+1}$ is passed to the next layer across $N$ layers.
%
However, this fixed, single-scale amplification applies a uniform frequency band to all variables, lacking the flexibility to encode heterogeneous spectral characteristics, where some fields vary smoothly while others contain sharp, high-frequency structures.
%
Recent frequency-adaptive methods~\cite{lindell2022bacon,zhang2023frequency} address this by predicting frequency shifts through auxiliary networks.
While effective, this reliance introduces parameter overhead that conflicts with the goal of compression, particularly under the multivariate setting.

To address this limitation, we introduce A$^2$-SIREN, a simple yet efficient mechanism that integrates a channel-wise learnable parameter ($\boldsymbol{\alpha}$).
Unlike these methods that depend on a separate prediction network, A$^2$-SIREN achieves frequency adaptation through a purely intrinsic mechanism.
%
Each layer is assigned a fixed base frequency $\omega_i \in [ 30, 150 ]$, and the learnable vector $\boldsymbol{\alpha}_i$ rescales it independently per channel.
%
A$^2$-SIREN is formulated as:
\begin{equation}
    \label{eq:A2-siren}
    \mathbf{x}_{i+1} = \mathtt{sin}( (\boldsymbol{\alpha}_i \odot \omega_i)\, \mathbf{h}_i(\mathbf{x}_i)), \quad \boldsymbol{\alpha}_i \in \mathbb{R}^{d},
\end{equation}
%
%
%
where $\odot$ scales the scalar base frequency $\omega_i$ channel-wise by $\boldsymbol{\alpha}_i \in \mathbb{R}^{d}$, yielding a per-channel frequency, and $d$ is the channel dimension.
Since only $\boldsymbol{\alpha}_i$ is learned while $\omega_i$ remains fixed, distinct variables can allocate independent frequency bandwidths within a single coordinate-based network.
Consequently, this intrinsic adaptability effectively captures physical features at negligible computational cost.

\subsection{SWAP Module}
%

The proposed SWAP module is a lightweight mechanism for efficiently encoding multivariate data, comprising three core components: a cross-variable attention layer, a path switching block, and layer pruning.

\subsubsection{Cross-variable attention layer}
As discussed, although multivariate scientific data exhibit inherent correlations, traditional shared networks~\cite{wurster2023adaptively,han2025dcinr,han2026moe} force variables to compete for the same parameters, causing severe feature interference.
Conversely, recent branched architectures~\cite{son2025mc} mitigate this by isolating variables, but their structural separation prevents the network from modeling explicit cross-variable interactions.
To overcome this limitation, we introduce a cross-variable attention layer that facilitates explicit interactions among all variables.
At a designated intermediate layer, this attention layer concatenates the independent branch features along the variable dimension into $\mathbf{X} = [\mathbf{x}^{(1)}, \mathbf{x}^{(2)}, \dots, \mathbf{x}^{(M)}]^\top \in \mathbb{R}^{M \times d}$, where $m$ and $M$ denote the variable index and the number of physical variables, respectively.
%
%
%
The query ($\mathbf{Q}$), key ($\mathbf{K}$), and value ($\mathbf{V}$) matrices are derived from $\mathbf{X}$ after layer normalization (LN) for stable optimization, and the cross-variable attention is formulated as:
%
%
\begin{align}
    \label{eq:cva}
        &\mathbf{X} = \mathtt{concat}(\mathbf{x}^{(1)}, \mathbf{x}^{(2)}, \dots, \mathbf{x}^{(M)}), \\
        &\mathbf{Q}, \mathbf{K}, \mathbf{V} = \text{Linear}_{Q, K, V}(\mathtt{LN}(\mathbf{X})), \\
        &\hat{\mathbf{X}} = \mathtt{softmax}\!\left( \frac{\mathbf{Q} \mathbf{K}^\top}{\sqrt{d_k}} \right) \mathbf{V}.
\end{align}
%
%
The attended output ($ \hat{\mathbf{X}}$) is residually added to $\mathbf{X}$ and injected into the next hidden layer ($\mathbf{h}_i$).
%
For instance, in SMR simulations, strongly coupled quantities such as temperature and velocity fields can exchange information through this layer, capturing their shared dynamics.
Thus, our approach harmonizes explicit multivariate interactions with the decoupled physical features essential for accurate scientific data modeling.

\input{_fig/fig_tex/fig3_pruning}

\subsubsection{Path switching block}
While a cross-variable attention layer facilitates explicit interactions, it implicitly assumes uniform correlation across all variables.
In reality, scientific data are heterogeneous: physical quantities may be strongly coupled or strictly independent.
Compounding this, attention aggregates features as a weighted average and thus acts as a low-pass filter~\cite{park2022vision,wang2022anti}. 
Such smoothing benefits coupled variables by reinforcing their shared dynamics, but blurs the sharp, high-frequency structures that characterize independent ones.
A cross-variable attention layer cannot serve both regimes simultaneously.
This motivates a mechanism that routes each variable based on whether it benefits from coupling or must preserve high-frequency detail.

To this end, we propose a path switching block with an attention path and a non-attention path, as described in~\figref{fig:pruning}.
%
To ensure stable optimization, we employ a sequential training strategy.
%
The non-attention path is first trained to establish a robust baseline of independent representations.
%
The attention path is then introduced and optimized while the non-attention path and shared layers remain frozen.
Upon completion, we apply a loss-driven path-selection mechanism.
%
For each variable $m$, we compute its converged per-variable attention-path loss $L_{\text{on}}^{(m)}$ and non-attention-path loss $L_{\text{off}}^{(m)}$.
We then define a binary switching mask $s^{(m)} \in \{0, 1\}$ as:
\begin{equation}
    s^{(m)} = \mathbb{I} \left( L_{\text{on}}^{(m)} \leq L_{\text{off}}^{(m)} \right),
\end{equation}
where $\mathbb{I}$ denotes the indicator function, yielding 1 if the condition is true and 0 otherwise.
During inference, the final optimal representation $\tilde{\mathbf{x}}^{(m)}$ for the $m$-th variable is deterministically computed as:
\begin{equation}
    \tilde{\mathbf{x}}^{(m)} = s^{(m)} \hat{\mathbf{x}}_{\text{attn}}^{(m)} + (1 - s^{(m)}) \hat{\mathbf{x}}_{\text{non-attn}}^{(m)},
\end{equation}
where $\hat{\mathbf{x}}_{\text{attn}}^{(m)}$ and $\hat{\mathbf{x}}_{\text{non-attn}}^{(m)}$ are the feature outputs of the respective paths, as detailed in~\figref{fig:pruning} (b).
%
%
%
This post-training selection ensures that the attention path is engaged only when coupling empirically benefits a variable, while variables that must retain high-frequency detail are preserved through the non-attention path.
%
The resulting routing is thus determined entirely by each variable's measured reconstruction loss, engaging cross-variable attention only where it is empirically beneficial and otherwise preserving the independent representation.
Our switching results are detailed in Appendix E. 

Although physical coupling can be spatially or temporally localized, we deliberately design the routing as a global, variable-level decision.
While localized routing could capture finer interactions, storing a dense coordinate-specific map incurs memory overhead.
Our block thus decouples variables at the macroscopic level while preserving the lightweight architecture for compression.

\subsubsection{Layer pruning}
After path switching, each variable activates only one path, leaving the unselected branches without any activation flow during the forward pass.
Since these inactive branches no longer contribute to the output, they become computationally redundant and can be safely removed.
Guided by the switching mask $s^{(m)}$, we therefore prune the network to retain only the active parameters.
Let $\Phi_{\text{non-attn}}^{(m)}$ and $\Phi_{\text{attn}}^{(m)}$ denote the parameters of the non-attention and attention paths for the $m$-th variable, where the attention path is further divided into pre-attention parameters $\Phi_{\text{pre}}^{(m)}$ (layers $l \le l_{\text{attn}}$) and post-attention parameters $\Phi_{\text{post}}^{(m)}$ (layers $l > l_{\text{attn}}$).
The retained parameter set $\tilde{\Phi}^{(m)}$ is formulated as:
\begin{equation}
    \tilde{\Phi}^{(m)} = 
        \begin{cases} \Phi_{\text{non-attn}}^{(m)} \cup \Phi_{\text{pre}}^{(m)}, & \text{if } s^{(m)} = 0,
        \\ \Phi_{\text{attn}}^{(m)}, & \text{if } s^{(m)} = 1.
        \end{cases}
\end{equation}
This conditional pruning retains different parameters depending on the selected path.
When $s^{(m)} = 0$, we prune the post-attention layers ($\Phi_{\text{post}}^{(m)}$) but preserve $\Phi_{\text{pre}}^{(m)}$.
As shown in \figref{fig:pruning} (c), this asymmetric retention is essential: it allows the $m$-th variable to still provide informative $\mathbf{K}$ and $\mathbf{V}$ for other actively attending variables.
When $s^{(m)} = 1$, the variable relies entirely on cross-variable interactions, so the independent non-attention path $\Phi_{\text{non-attn}}^{(m)}$ is discarded without any loss of essential features (\figref{fig:pruning} (d)).
Overall, this targeted pruning removes redundant operations, yielding a highly lightweight model while strictly preserving accuracy.

\subsection{Training Strategy and Loss Function}
To ensure stable optimization, we adopt a two-stage strategy.
For any output $\mathbf{\hat{x}}$ and ground truth (GT) $\mathbf{x}_{\text{gt}}$, the mean squared error (MSE) loss is:
%
\begin{equation}
    L(\hat{\mathbf{x}}) = \frac{1}{M} \sum_{m=1}^{M} \left\| \hat{\mathbf{x}}^{(m)} - \mathbf{x}_{\text{gt}}^{(m)} \right\|_2^2.
\end{equation}

\noindent\textbf{Stage 1 (non-attention path).}
We first establish a robust baseline of independent physical representations by optimizing the feature extractor and the non-attention path:
\begin{equation}
    \min_{\Phi_{\text{sh}}, \Phi_{\text{non-attn}}} L(\hat{\mathbf{x}}_{\text{non-attn}}),
\end{equation}
where $\Phi_{\text{sh}}$ denotes the parameters of shared MLP layers.

\noindent\textbf{Stage 2 (attention path).}
%
Once the non-attention path converges, we initialize the attention path with its weights ($\Phi_{\text{attn}} \leftarrow \Phi_{\text{non-attn}}$) and randomly initialize the new linear projections.
We then freeze the baseline parameters ($\Phi_{\text{sh}}$ and $\Phi_{\text{non-attn}}$) and optimize only the attention path:
\begin{equation}
    \min_{\Phi_{\text{attn}}} L(\hat{\mathbf{x}}_{\text{attn}}).
\end{equation}
%
Freezing the baseline lets the network explore cross-variable interactions without corrupting the learned independent representations.

\subsection{Quantization}
Layer pruning reduces structural redundancy, but the model size is ultimately governed by parameter precision.
To achieve a lightweight model, we apply post-training quantization as the final compression step, inspired by COIN~\cite{dupont2021coin}.
The network is trained in 32-bit floating-point precision for stable gradient updates.
For storage, only the parameters are quantized to 16-bit, while activations are cast at inference time for efficiency.
Since neural representations are largely resilient to reduced precision~\cite{micikevicius2017mixed}, this halves the storage budget, $M(\tilde{\Phi}_{q}) \approx \frac{1}{2} M(\tilde{\Phi})$, where $M(\cdot)$ denotes the overall model size, while incurring minor loss in reconstruction quality. 
%
As a result, SWAP-INR effectively doubles its compression ratio at nearly imperceptible cost.

\input{_tab/Tab1_dataset}


%% file: _fig/fig_tex/fig3_pruning.tex
\begin{figure}[!ht]
    \centering
    \includegraphics[width=0.99\linewidth]{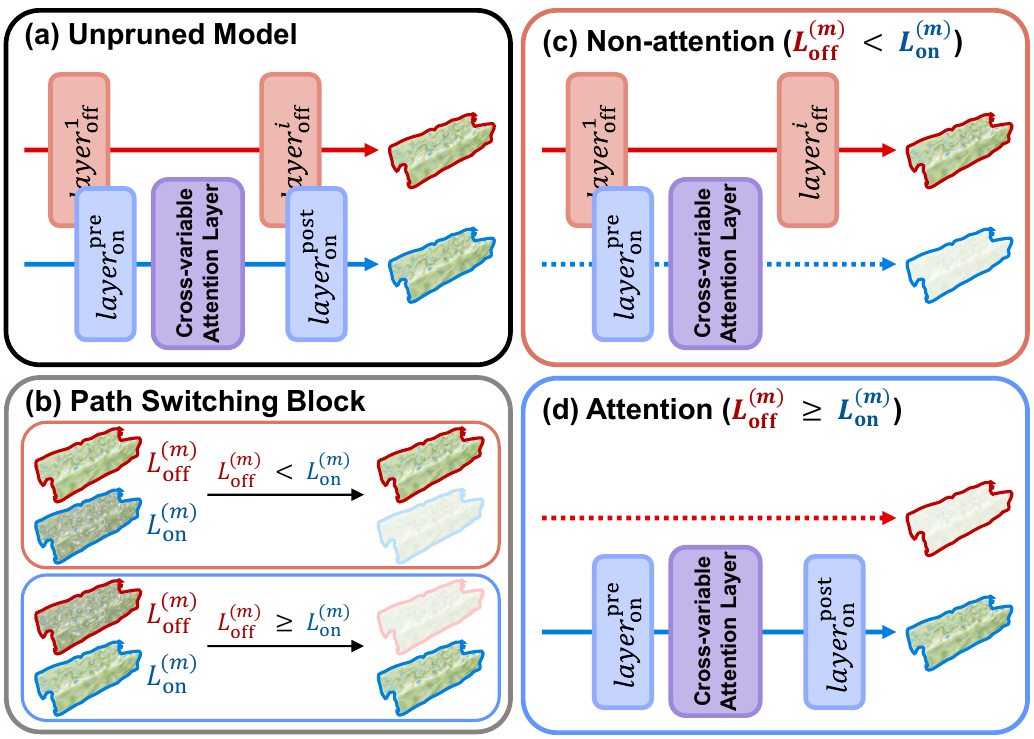}
    \caption{
    Illustration of the variable-driven switchable mechanism and targeted layer pruning.
    (a) The unpruned model features both non-attention (red) and attention (blue) paths.
    %
    (b) Path switching block
    evaluates $L^{(m)}_\mathrm{off}$ and $L^{(m)}_\mathrm{on}$ to deterministically route the $m$-th variable.
    (c) When the non-attention path is selected ($L^{(m)}_\mathrm{off} < L^{(m)}_\mathrm{on}$), post-attention layers are pruned while base layers are retained to serve as $\mathbf{K}, \mathbf{V}$.
    (d) Conversely, when the attention path is selected ($L^{(m)}_\mathrm{off} \ge L^{(m)}_\mathrm{on}$), the structurally redundant non-attention baseline path is eliminated.}
    \label{fig:pruning}
\end{figure}

%% file: _tab/Tab1_dataset.tex
\begin{table}[htbp]
\caption{
Experimental datasets. $\mathbf{P}$ and $t$ are the \# of spatial coordinates per timestep and the \# of timesteps, respectively. 
}
\label{tab:dataset}
\centering
\resizebox{0.99\linewidth}{!}{
\begin{tabular}{c|c|cc}
\Xhline{2\arrayrulewidth}
Grid type & Dataset & \# of points ($\mathbf{P}$ $\times$ $t$) & \# of variables\\ \hline \hline
\multirow{3}{*}{Unstructured} 
& ROCOM~\cite{cho2019numerical}   & 12,440,366 $\times$ 5     & 8\\  
& CUPID                           & 1,279,375  $\times$ 30    & 11\\
& Synthetic                       & 5,253,750  $\times$ 30    & 4\\
\hline
\multirow{3}{*}{Structured}
& SMR Simulation                         & 1,048,576 $\times$ 10       & 8 \\
& Combustion~\cite{hawkes2007scalar}     & 648,000   $\times$ 25       & 4 \\
& Ionization~\cite{whalen2008ionization} & 576,600   $\times$ 50       & 4 \\
\Xhline{2\arrayrulewidth}
\end{tabular}
}
\end{table}

%% file: _sec/4_exp.tex
\section{Experiments}
\label{sec:exp}

\input{_tab/Tab3_main_exp}

\subsection{Experiment Setup}
\noindent\textbf{Datasets.} Our experiments were conducted using a comprehensive suite of multivariate scientific datasets.
The datasets are categorized by their spatial topology: ROCOM~\cite{cho2019numerical}, CUPID, and Synthetic datasets are SMR data defined on unstructured tetrahedral meshes.
%
In contrast, the SMR simulation dataset (provided by our collaborator) and the publicly available Combustion~\cite{hawkes2007scalar} and Ionization~\cite{whalen2008ionization} datasets are defined on structured 3D rectilinear grids.
For computational feasibility, the Combustion and Ionization datasets were spatiotemporally downsampled by a factor of 4.
The details of each dataset are summarized in \Tabref{tab:dataset}.



\noindent\textbf{Implementation details.}
SWAP-INR comprises a shared base network (2 layers, 128 channels) and variable-specific branches (6 layers), each splitting into a non-attention and an attention path with the cross-variable attention layer applied after the third branched layer.
The per-channel scalar $\boldsymbol{\alpha}$ is uniformly initialized to 1.0, and the number of feature channels per branch is set by dataset complexity: 64 for ROCOM and Synthetic, 56 for CUPID, and 48 for SMR Simulation, Combustion, and Ionization.
Each physical variable is independently normalized to $[-1, 1]$.
We train all models in PyTorch 2.1.0 on a single NVIDIA RTX 3090 GPU using Adam (learning rate $2 \times 10^{-5}$, batch size 100,000), with early stopping after 20 epochs without loss improvement.
For evaluation, we report peak signal-to-noise ratio (PSNR), normalized root-mean-squared error (NRMSE), the coefficient of determination ($R^2$), compression ratio (CR), and model size in KB.
Note that CR is defined as raw data size over total stored network parameters; mesh topology is excluded as INRs represent fields grid-agnostically without encoding the evaluation grid itself.

%
%

\input{_fig/fig_tex/fig6_main_comparison}

\subsection{Results and Evaluation}
We compare against a broad range of methods: INR-based approaches SIREN~\cite{sitzmann2020implicit}, NeurComp~\cite{lu2021compressive}, CoordNet~\cite{han2022coordnet}, Instant-NGP~\cite{muller2022instant}, fV-SRN~\cite{weiss2022fast}, WIRE~\cite{saragadam2023wire}, AFNet~\cite{zhang2023frequency}, APMGSRN~\cite{wurster2023adaptively}, SL$^2$A-INR~\cite{rezaeian2025sl2a}, and MC-INR~\cite{son2025mc}, as well as the conventional lossy compressors SZ3~\cite{liang2022sz3}, ZFP~\cite{lindstrom2014fixed}, and TTHRESH~\cite{ballester2019tthresh}. 
Further details on these methods and the matched PSNR comparison are provided in Appendix A 
and Appendix B. 

\subsubsection{Quantitative results at matched CR}

\Tabref{tab:main_exp} presents a quantitative comparison against previous INR-based methods at matched CR.
Note that fV-SRN and APMGSRN do not support time-varying data; we evaluated them per time step and averaged the results.
SWAP-INR achieves SOTA performance on the SMR datasets and competitive results on the public Combustion and Ionization datasets.
This confirms that explicitly modeling cross-variable interactions, rather than naively sharing or fully isolating them, enhances reconstruction fidelity across diverse data.
The lower performance here stems from the low-pass nature of the attention path~\cite{park2022vision,wang2022anti}, analyzed in Appendix C. 
%
SWAP-INR maintains a competitive fidelity-efficiency trade-off among methods of comparable speed (Appendix D), and remains SOTA on the in-house SMR datasets under 16-bit quantization (Appendix F).

\subsubsection{Qualitative results at matched CR}
%
\Figref{fig:main_comparison} compares SWAP-INR with CoordNet, fV-SRN, SL$^2$A-INR, and MC-INR via error maps. 
%
SWAP-INR consistently achieves the lowest error across the datasets, and the error maps expose a distinct failure mode behind each baseline.
%
The CoordNet shared layer causes cross-variable interference that spreads error broadly.
fV-SRN reconstructs each time step independently, accumulating temporally inconsistent errors on CUPID and Ionization.
SL$^2$A-INR, lacking explicit cross-variable modeling, retains visibly higher residual error across coupled regions.
MC-INR clusters, when constrained to the target CR, leave high-error blocks at their boundaries (ROCOM, Combustion, Ionization).
In contrast, SWAP-INR couples variables that benefit from interaction while preserving high-frequency details in the others.
This yields consistently low reconstruction error and faithful recovery of fine physical structures critical for scientific visualization.
These results demonstrate the quantitative and qualitative advantages of SWAP-INR.

\subsection{Ablation Study}
\label{sec:ablation}

We conduct ablations on the ROCOM, SMR Simulation, and Ionization datasets to validate three components: (1) the adaptability amplification factor $\boldsymbol{\alpha}$; (2) the elements of the SWAP module, and (3) the quantization process.
\\

\input{_tab/Tab5_A2}

\noindent\textbf{Adaptability amplification factor.}
\Tabref{tab:ablation_alpha} reports the effect of $\boldsymbol{\alpha}$.
Applying it improves PSNR by 1.12 dB on ROCOM, 0.38 dB on SMR Simulation, and 0.16 dB on Ionization, confirming that channel-wise frequency modulation better captures physical details. 
Notably, this improvement requires only an additional 22 KB, demonstrating that $\boldsymbol{\alpha}$ provides a simple yet effective mechanism for balancing accuracy and memory efficiency.


\input{_tab/Tab6_module}

\noindent\textbf{SWAP module components.}
\Tabref{tab:ablation_SWAP} validates each component within the SWAP module.
The cross-variable attention layer increases PSNR by 1.03~dB (ROCOM), 0.89~dB (SMR Simulation), and 2.47~dB (Ionization) by explicitly capturing multivariate interactions.
%
The path switching block yields further gains by routing each variable to its lower-loss path, isolating variables that do not benefit from coupling to prevent feature interference, with full routing results in Appendix E. 
Layer pruning reduces the model size by 41\% (988 to 580 KB) on ROCOM, 27\% (674 to 491 KB) on SMR Simulation, and 41\% (377 to 224 KB) on Ionization, with no drop in reconstruction quality.
These results demonstrate that every component is integral to maximizing both the fidelity and compactness of SWAP-INR.

\input{_tab/Tab7_quantization}


\noindent\textbf{Quantization.}
\Tabref{tab:ablation_quan} reports the effect of quantization across 32-, 16-, and 8-bit precisions.
Reducing the precision from 32-bit to 16-bit approximately halves the model size while preserving reconstruction quality.
%
Crucially, Appendix F demonstrates that this reduction yields nearly imperceptible visual differences.
%
Under this uniform 16-bit setup, SWAP-INR outperforms all baselines except on Combustion, confirming that our overall advantage stems from architectural representation rather than reduced precision.
Conversely, 8-bit quantization severely degrades performance across all datasets. 
Thus, 16-bit precision serves as the optimal operating point, effectively doubling the CR while preserving high reconstruction fidelity.

%% file: _tab/Tab3_main_exp.tex
\begin{table*}[!ht]
\centering
\caption{
    Quantitative comparison of compression results with previous methods at matched CR.
    The $\ast$ denotes that the reported performance is averaged across independent time steps.
    Baselines are stored at 32-bit precision, while SWAP-INR uses 16-bit.
    }
\label{tab:main_exp}
\resizebox{0.99\linewidth}{!}{
\begin{tabular}{c|c|ccccc|c|c|ccccc}
\Xhline{2\arrayrulewidth}
Dataset & Method & PSNR$\uparrow$ & NRMSE$\downarrow$ & $R^2\uparrow$ & CR$\uparrow$ & Size(KB)$\downarrow$
& Dataset & Method & PSNR$\uparrow$ & NRMSE$\downarrow$ & $R^2\uparrow$ & CR$\uparrow$ & Size(KB)$\downarrow$ \\
\hline
\hline

\multirow{11}{*}{ROCOM}
& SIREN              & 42.34 & 0.0114 & 0.9323  & 3,311 & 587             
& \multirow{11}{*}{CUPID}
& SIREN              & 21.46 & 0.1657 & 0.5585  & 2,634 & 626   \\        
& NeurComp           & 42.66 & 0.0105 & 0.9398  & 3,346  & 581            
& & NeurComp         & 24.63 & 0.0850 & 0.7002  & 2,626 & 628  \\         
& CoordNet           & 43.49 & 0.0105 & 0.9392  & 3,351 & 580             
& & CoordNet         & 23.87 & 0.0914 & 0.6540  & 2,609 & 632 \\          
& Instant-NGP        & 30.28 & 0.0356 & 0.4391  & 3,346 & 581             
& & Instant-NGP      & 19.61 & 0.1791 & 0.3948  & 2,561 & 644 \\          
& fV-SRN$^\ast$      & 40.24 & 0.0125 & 0.8764  & 3,351 & 580             
& & fV-SRN$^\ast$    & 13.99 & 0.2362 & -13.4032& 2,618 & 630 \\          
& WIRE               & 25.09 & 0.0966 & -0.0004 & 3,351 & 580             
& & WIRE             & 16.92 & 0.2595 & 0.0000  & 2,630 & 627 \\          
& AFNet              & \underline{44.19} & \underline{0.0094} & \underline{0.9480}  & 3,351 & 580             
& & AFNet            & \underline{48.38} & \underline{0.0044} & \underline{0.9695}  & 2,622 & 629 \\          
& APMGSRN$^\ast$     & 38.21 & 0.0171 & 0.8383  & 3,351 & 580             
& & APMGSRN$^\ast$   & 19.25 & 0.1789 & -0.1587 & 2,618 & 630 \\          
& SL$^2$A-INR        &43.39  & 0.0113 & 0.9258  & 3,334 & 583             
& & SL$^2$A-INR      & 46.77 & 0.0051 & 0.9679  & 2,626 & 628 \\          
& MC-INR             & 33.61 & 0.0255 & 0.5857  & 3,328 & 584             
& & MC-INR           & 46.89 & 0.0086 & 0.9579  & 2,440 & 676 \\   
& \textbf{Ours}  & \textbf{47.41} & \textbf{0.0073} & \textbf{0.9782} & 3,351  & 580        
& & \textbf{Ours}& \textbf{50.29} & \textbf{0.0035} & \textbf{0.9904} & 2,634  & 626  \\    

\hline
\hline

\multirow{11}{*}{Synthetic}
& SIREN              & 70.31 & \underline{0.0003} & 0.9245 & 7,793 & 316  
& \multirow{11}{*}{\begin{tabular}{@{}c@{}}SMR \\ Simulation\end{tabular}}
& SIREN              & 33.71 & 0.0277 & 0.2851  & 665   & 493 \\          
& NeurComp           & 68.72 & 0.0004 & 0.8830  & 7,868 & 313             
& & NeurComp         & 33.15 & 0.0297 & 0.1884  & 665   & 493 \\          
& CoordNet           & 72.14 & \underline{0.0003} & \textbf{0.9755} & 7,843  & 314   
& & CoordNet         & 34.03 & 0.0267 & 0.3411 & 666  & 492 \\            
& Instant-NGP        & 35.22 & 0.0776 & 0.0280  & 7,601  & 324            
& & Instant-NGP      & 30.16 & 0.0850 & 0.0356  & 635  & 516  \\          
& fV-SRN$^\ast$      & 64.10 & 0.0007 & 0.8038  & 7,463  & 330            
& & fV-SRN$^\ast$    & 29.60 & 0.0472 & -3.1898 & 655  & 500  \\          
& WIRE               & 34.90 & 0.0802 & -0.0470 & 7,868 & 313             
& & WIRE             & 29.98 & 0.0911 & -0.0024 & 663   & 494 \\          
& AFNet              & \underline{74.60} &\textbf{0.0002} & 0.9657  & 7,793 & 316             
& & AFNet            & \underline{35.80} & \underline{0.0190} & \underline{0.3564} & 666  & 492 \\       
& APMGSRN$^\ast$     & 60.93 & 0.0011 & 0.7438  & 7,463 & 330             
& & APMGSRN$^\ast$   & 30.51 & 0.0534 & -0.6753 & 655 & 500  \\           
& SL$^2$A-INR        & 73.00 & \textbf{0.0002} & 0.9338  & 7,793 & 316             
& & SL$^2$A-INR      & 35.10 & 0.0210 & 0.3108  & 667   & 491 \\          
& MC-INR             & 66.69 & 0.0006 & 0.9616  & 7,696 & 320             
& & MC-INR           & 33.92 & 0.0259 & 0.2826  & 655 & 500  \\           
& \textbf{Ours}  & \textbf{75.00} & \textbf{0.0002} & \underline{0.9736} & 7,868 & 313 
& & \textbf{Ours}& \textbf{38.17} & \textbf{0.0153} & \textbf{0.4494} & 667 & 491 \\ 

\hline
\hline

\multirow{11}{*}{Combustion}
& SIREN               & 30.62 & 0.0296 & 0.8967  & 989 & 256              
& \multirow{11}{*}{Ionization}
& SIREN               & 43.12 & 0.0078 & 0.9664  & 2,011 & 224   \\       
& NeurComp            & \textbf{32.22} & \textbf{0.0245} & \textbf{0.9140} & 997 & 254     
& & NeurComp          & 45.67 & 0.0056 & 0.9777  & 2,002 & 225   \\       
& CoordNet            & 31.09 & 0.0280 & 0.9014  & 981   & 258            
& & CoordNet          & 39.28 & 0.0135 & 0.9451  & 1,993 & 226  \\        
& Instant-NGP         & 20.86 & 0.1033 & 0.4767  & 977   & 259            
& & Instant-NGP       & 21.93 & 0.1988 & 0.1837  & 1,984 & 227 \\         
& fV-SRN$^\ast$       & 23.53 & 0.0726 & 0.6768  & 920   & 275            
& & fV-SRN$^\ast$     & 22.76 & 0.1039 & -0.6598 & 1,802 & 250 \\         
& WIRE                & 17.75 & 0.1615 & 0.0000  & 985   & 257            
& & WIRE              & 20.94 & 0.2469 & -0.0002 & 2,011 & 224 \\         
& AFNet               & 28.33 & 0.0392 & 0.8643  & 997   & 254            
& & AFNet             & \textbf{50.62} & \textbf{0.0030} & \underline{0.9899}  & 2,011 & 224 \\         
& APMGSRN$^\ast$      & 21.52 & 0.0951 & 0.5493  & 920   & 275            
& & APMGSRN$^\ast$    & 18.02 & 0.1439 & -9.2996 & 1,802 & 250 \\         
& SL$^2$A-INR         & 26.02 & 0.0525 & 0.8025  & 981   & 258            
& & SL$^2$A-INR       & 42.64 & 0.0078 & 0.9496  & 2,002 & 225 \\         
& MC-INR              & 18.05 & 0.1502 & -0.0065 & 989   & 256            
& & MC-INR            & 20.93 & 0.1686 & -0.3931 & 1,909 & 236  \\        
& \textbf{Ours}   & \underline{31.17} & \underline{0.0278} & \underline{0.9116} & 997 & 254   
& & \textbf{Ours} & \underline{50.25} & \underline{0.0032} & \textbf{0.9905}  & 2,011  & 224  \\        

\Xhline{2\arrayrulewidth}
\end{tabular}
}
\end{table*}

%% file: _fig/fig_tex/fig6_main_comparison.tex
\begin{figure*}[!ht]
    \centering
    \includegraphics[width=0.97\linewidth]{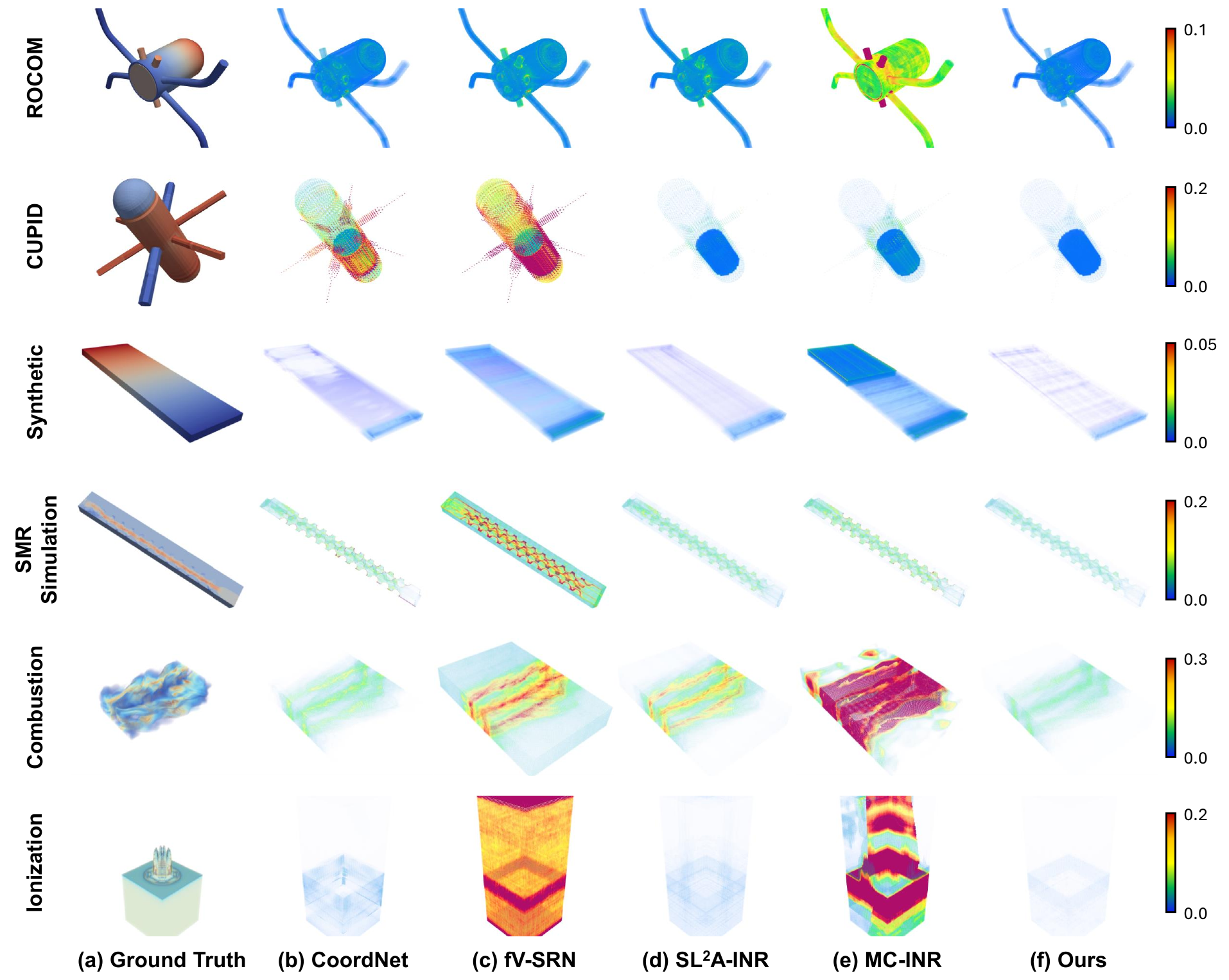}
    \caption{
    Visualization of error maps at matched compression ratios. 
    The error map represents the mean absolute error 
    averaged across all time steps and variables (blue: low, red: high).
    GT is visualized for a specific snapshot (time and variable).
    ROCOM, CUPID, Synthetic, and SMR Simulation are rendered via tetrahedral meshes, whereas Combustion and Ionization are shown as regular volumes.
    }
    \label{fig:main_comparison}
\end{figure*}

%% file: _tab/Tab5_A2.tex
\begin{table}[!htbp]
\centering
\caption{Ablation on adaptability amplification factor, $\boldsymbol{\alpha}$.}
\label{tab:ablation_alpha}

\resizebox{0.85\linewidth}{!}{
\begin{tabular}{c|c|ccc}
\Xhline{2\arrayrulewidth}
Dataset & Method & PSNR$\uparrow$ & NRMSE$\downarrow$ & $R^2\uparrow$ \\
\hline
\hline

\multirow{2}{*}{ROCOM}
& w/o $\boldsymbol{\alpha}$ & 46.29 & 0.0082 & 0.9665 \\
& w/ $\boldsymbol{\alpha}$  & \textbf{47.41} & \textbf{0.0073} & \textbf{0.9782} \\
\hline

SMR & w/o $\boldsymbol{\alpha}$ & 37.79          & 0.0158          & 0.4195 \\
Simulation & w/ $\boldsymbol{\alpha}$  & \textbf{38.17} & \textbf{0.0153} & \textbf{0.4494} \\

\hline
\multirow{2}{*}{Ionization}
& w/o $\boldsymbol{\alpha}$ & 50.09 & 0.0033 & 0.9899 \\
& w/ $\boldsymbol{\alpha}$  & \textbf{50.25} & \textbf{0.0032} & \textbf{0.9905} \\

\Xhline{2\arrayrulewidth}

\end{tabular}
}
\end{table}

%% file: _tab/Tab6_module.tex
\begin{table}[!htbp]
\centering
\caption{
Ablation on the SWAP module. We evaluate the individual contributions of its core components: (a) cross-variable attention layer, (b) path switching block, and (c) layer pruning.
}
\label{tab:ablation_SWAP}

\resizebox{0.99\linewidth}{!}{
\begin{tabular}{c|ccc|ccccc}
\Xhline{2\arrayrulewidth}
Dataset & (a) & (b) & (c) & PSNR$\uparrow$ & NRMSE$\downarrow$ & $R^2\uparrow$ & CR$\uparrow$ & Size(KB)$\downarrow$\\
\hline
\hline

\multirow{4}{*}{ROCOM}
&            &            &            & 46.20 & 0.0080 & 0.9742 & 1,967  & 988 \\
& \checkmark &            &            & 47.23 & 0.0073 & 0.9782 & 1,967  & 988 \\
& \checkmark & \checkmark &            & 47.41 & 0.0073 & 0.9782 & 1,967  & 988 \\
& \checkmark & \checkmark & \checkmark & 47.41 & 0.0073 & 0.9782 & 3,351  & 580 \\

\hline

\multirow{4}{*}{\begin{tabular}{@{}c@{}}SMR \\ Simulation\end{tabular}}
&            &            &             & 37.25 & 0.0161 & 0.4448 & 486 & 674 \\
& \checkmark &            &             & 38.14 & 0.0154 & 0.4395 & 486 & 674 \\
& \checkmark & \checkmark &             & 38.17 & 0.0153 & 0.4494 & 486 & 674 \\
& \checkmark & \checkmark & \checkmark  & 38.17 & 0.0153 & 0.4494 & 667 & 491 \\

\hline

\multirow{4}{*}{Ionization}
&            &            &             & 47.78 & 0.0044 & 0.9854 & 1,195 & 377 \\
& \checkmark &            &             & 50.25 & 0.0032 & 0.9905 & 1,195 & 377 \\
& \checkmark & \checkmark &             & 50.25 & 0.0032 & 0.9905 & 1,195 & 377 \\
& \checkmark & \checkmark & \checkmark  & 50.25 & 0.0032 & 0.9905 & 2,011 & 224 \\

\Xhline{2\arrayrulewidth}
\end{tabular}
}

\end{table}

%% file: _tab/Tab7_quantization.tex
\begin{table}[!htbp]
\centering
\caption{Ablation on model bit-widths.}
\label{tab:ablation_quan}

\resizebox{0.99\linewidth}{!}{
\begin{tabular}{c|c|cccc}
\Xhline{2\arrayrulewidth}
Dataset & Bit-widths & PSNR$\uparrow$ & NRMSE$\downarrow$ & $R^2\uparrow$ & Size(KB)$\downarrow$ \\
\hline
\hline

\multirow{3}{*}{ROCOM}
& 32-bit & 47.89 & 0.0072 & 0.9788 & 1,101 \\
& 16-bit & 47.41 & 0.0073 & 0.9782 & 580 \\
& 8-bit & 24.41 & 0.0642 & -6.5984 & 319 \\
\hline

\multirow{3}{*}{CUPID}
& 32-bit & 50.91 & 0.0033 & 0.9912 & 1,167 \\
& 16-bit & 50.29 & 0.0035 & 0.9904 & 626  \\
& 8-bit & 16.19 & 0.1707 & -22.2139 & 351  \\
\hline

\multirow{3}{*}{Synthetic}
& 32-bit & 78.10 & 0.0001 & 0.9769 & 596 \\
& 16-bit & 75.00 & 0.0002 & 0.9736 & 313 \\
& 8-bit & 40.48 & 0.0117 & -500.3126 & 172 \\
\hline

\multirow{3}{*}{\begin{tabular}{@{}c@{}}SMR \\ Simulation\end{tabular}}
& 32-bit & 38.51 & 0.0152 & 0.4502 & 919 \\
& 16-bit & 38.17 & 0.0153 & 0.4494 & 491 \\
& 8-bit & 22.20 & 0.0852 & -92.1838 & 285 \\
\hline

\multirow{3}{*}{Combustion}
& 32-bit & 31.20 & 0.0277 & 0.9119 & 478 \\
& 16-bit & 31.17 & 0.0278 & 0.9116 & 254 \\
& 8-bit & 22.84 & 0.0771 & 0.6117 & 146 \\
\hline

\multirow{3}{*}{Ionization}
& 32-bit & 52.52 & 0.0024 & 0.9933 & 421 \\
& 16-bit & 50.25 & 0.0032 & 0.9905 & 224 \\
& 8-bit & 20.25 & 0.1044 & -33.0035 & 129 \\

\Xhline{2\arrayrulewidth}

\end{tabular}
}
\end{table}

%% file: _sec/5_con.tex
\section{Conclusion}

\label{sec:conclusion}

We presented SWAP-INR, an INR for high-fidelity compression of time-varying multivariate scientific data on both structured and unstructured grids.
%
At its core, the SWAP module routes each variable to an attention or non-attention path by a deterministic, loss-driven criterion and prunes the inactive branches.
Together with the channel-wise frequency modulation of A$^2$-SIREN, this models cross-variable interactions within a minimal parameter footprint.
Combined with 16-bit quantization, SWAP-INR achieves SOTA performance on the SMR datasets and competitive results on two public datasets at matched compression ratios, while maintaining inference cost on par with lightweight INR baselines.
In practice, this lets engineers store, transmit, and visualize massive simulations on complex geometries within tight HPC budgets, addressing a bottleneck in safety-critical workflows like SMR analysis.

For future work, frequency-adaptive attention could mitigate over-smoothing in high-frequency fields, and extending evaluation to downstream tasks such as derived-quantity analysis could further demonstrate scientific utility.

%% file: _sec/6_sup.tex
\section*{Appendix}
\label{sec:sup}


\noindent In this appendix, we present additional implementation details, results at matched PSNR, an extended discussion of the low-pass filter of attention and \textit{hot corner} cases,  path-switching results, training-time comparisons, additional results for 16-bit quantization, and CR-PSNR curves.

\appendix
\section{Additional Implementation Details}

\input{_tab/Tab2_comparison_model}
\label{sec:comparison_method}
\Tabref{tab:methods} summarizes the structural properties of mainstream INR-based and traditional compressors along four axes central to multivariate scientific data: native multivariate support (Multi.), grid-agnostic operation (Grid-agn.), interference-free representation (No-Interf.), and explicit cross-variable correlation (Corr.).
Conventional lossy compressors (ZFP, TTHRESH, SZ3) handle multiple variables but are tied to discrete grids and cannot model cross-variable correlations.
Most INR-based methods, in contrast, target single-variable fields; even those supporting multiple variables either share parameters across them, incurring feature interference, or isolate them entirely, discarding correlations.
As the table shows, no existing method satisfies all four properties at once, whereas SWAP-INR is the only one to do so, directly reflecting our design goals.

For evaluation, we use the official source code for every method that provides it and exclude those without a public implementation, adapting several baselines to our setting.
Specifically, the gradient-based regularization of NeurComp is modified to operate on unstructured data, and since the ZFP and TTHRESH require structured grids, we convert the unstructured datasets accordingly.
%

\section{Results at matched PSNR}
\label{sec:match_PSNR}
\input{_tab/Tab4_compressor}
\noindent\textbf{Comparison at matched PSNR.}
\Tabref{tab:compression} evaluates our method against conventional lossy compressors at matched PSNR.
Since ZFP and TTHRESH operate exclusively on structured grids, we convert the ROCOM~\cite{cho2019numerical}, CUPID, and Synthetic datasets into structured formats for fair evaluation.
SWAP-INR achieves substantially higher CR and smaller model size on most datasets, indicating that our continuous representation with targeted pruning and quantization is more compact than conventional methods.
One exception is the Synthetic dataset, where TTHRESH yields a higher CR due to a localized high-intensity region (the \textit{hot corner} phenomenon), detailed in Appendix~\ref{sec:discussion}.
SWAP-INR maintains its practical superiority across all other datasets.

\input{_fig/fig_tex/fig7_compressor_comparison}
\noindent\textbf{Visualization at matched PSNR}
\figref{fig:compressor_comparison} compares SWAP-INR with conventional lossy compressors.
We evaluate only the structured datasets, as converting unstructured data into regular volumes introduces resampling artifacts that confound visual comparison.
Although all methods are constrained to the same PSNR, the lossy compressors exhibit pronounced noise and visual artifacts.
SWAP-INR instead suppresses noise and favors smooth reconstructions via learning-based continuous INRs and a cross-variable attention layer, yielding the highest visual fidelity, particularly valuable in scientific visualization.
Together with the smaller memory budget in \tabref{tab:compression}, these results demonstrate the quantitative and qualitative superiority of SWAP-INR.

\section{Discussion}
\label{sec:discussion}
\input{_fig/fig_tex/fig4_gradient}

\noindent\textbf{Combustion: low-pass filter of attention.}
On Combustion, SWAP-INR yields an overall PSNR 1.05~dB below NeurComp.
This gap originates from the attention operation, which aggregates features as a weighted average and thus acts as a low-pass filter, smoothing sharp structures while suppressing noise~\cite{park2022vision,wang2022anti}.
As shown in~\figref{fig:vs_neurcomp} (top), variables on the attention path exhibit slight blurring but fewer structural artifacts than NeurComp.
%
Conversely, variables on the non-attention path (bottom) avoid the low-pass effect of attention and, aided by A$^2$-SIREN, preserve high-frequency detail, surpassing NeurComp by 0.14~dB.
Crucially, this lower PSNR is not a failure but a design choice: SWAP-INR routes each variable to balance fidelity against sharpness, trading a marginal PSNR drop for robust reconstructions suited to scientific visualization.

\input{_fig/fig_tex/fig_ionization}

\noindent\textbf{Ionization: the dual benefit of attention.}
On Ionization, SWAP-INR is 0.37~dB below AFNet in PSNR, yet the two differ in visual quality.
As shown in~\figref{fig:vs_afnet}, the attention path offers two complementary benefits depending on each variable's role.
The top row shows that AFNet introduces noticeable structural artifacts throughout the volume, whereas the low-pass nature of attention suppresses them, yielding clean, artifact-free reconstructions.
The bottom row shows that attention introduces slight blurring as a weighted average, yet recovers the underlying structure more faithfully than AFNet by exploiting cross-variable correlation.
Across both regimes, the attention path thus suppresses artifacts while preserving physical structure, precisely the reliability demanded by scientific visualization, where such artifacts can mislead downstream analysis.
SWAP-INR therefore turns a marginal PSNR gap into a tangible gain in visual fidelity, underscoring that numerical scores alone do not capture reconstruction quality.

\input{_fig/fig_tex/fig5_hotcorner}
\noindent\textbf{Synthetic: the \textit{hot corner} phenomenon.}
On Synthetic, TTHRESH attains a higher compression ratio than SWAP-INR, the only dataset where a conventional compressor surpasses our method (\Tabref{tab:compression}).
As visualized in~\figref{fig:hotcorner}, this stems from the \textit{hot corner} phenomenon: under HOSVD, the structural energy of the core tensor concentrates almost entirely in a single corner, leaving the remainder near zero.
Such extreme, localized sparsity is the ideal regime for transform-based compressors, as TTHRESH directly discards the resulting zero coefficients.
Coordinate-based INRs, by contrast, distribute capacity smoothly over the domain and cannot concentrate it so asymmetrically, a manifestation of their well-known spatial-spectral bias.
We regard this as a boundary condition specific to such degenerate distributions rather than a general limitation, where SWAP-INR otherwise consistently outperforms.

\section{Training time}
\label{sec:training_time}
\Tabref{tab:time} compares the per-epoch time $t$ (sec), the total training time $t$ (hr), and the full-volume inference time across all datasets, where every model is trained until convergence (no loss improvement over 20 consecutive epochs).
Despite its cross-variable attention layer and adaptive frequency modulation, SWAP-INR maintains efficient per-epoch computation, comparable to lightweight baselines such as SIREN, NeurComp, and CoordNet.
Total training time instead depends on how long each model keeps improving before convergence.
Note that fV-SRN and APMGSRN cannot handle time-varying data; they train a separate model per time step, inflating their totals.
SWAP-INR, by comparison, converges quickly and stays among the most efficient on all datasets except Combustion.
There, the loss drops sharply to about $3.6\times10^{-3}$ by the $20$-hour mark and then declines marginally to $3.2\times10^{-3}$ near $50$ hours, reflecting continued refinement of fine details rather than wasted computation.

SWAP-INR's inference time is on par with lightweight baselines such as SIREN, NeurComp, and CoordNet, and faster than per-timestep methods (fV-SRN, APMGSRN) that decode each step separately.
It matches these baselines because pruning removes the inactive path for each variable, so only a single path is evaluated at inference, keeping the per-query cost close to a standard coordinate-based network.
Crucially, SWAP-INR achieves the best compression-fidelity trade-off, attaining the highest PSNR at matched CR among methods of similar speed.

\input{_tab/Sup_Tab_time}

\section{Path Switching Results}
Tables~\ref{tab:routing_ROCOM} through~\ref{tab:routing_ionization} report the path-switching results for all six datasets: ROCOM~\cite{cho2019numerical}, CUPID, Synthetic, SMR Simulation, Combustion~\cite{hawkes2007scalar}, and Ionization~\cite{whalen2008ionization}.
For each dataset, we list the path selected for every variable along with its reconstruction PSNR.
The routing is determined entirely by the deterministic, loss-driven criterion, without any hand-specified correlation structure or manual tuning.
The resulting assignments are non-trivial and dataset-dependent: rather than collapsing to a single path, variables split between the attention and non-attention paths in proportions that vary across datasets.
Critically, this routing is not merely descriptive but functionally validated.
The SWAP module ablation shows that enabling switching consistently improves reconstruction fidelity.
This confirms that the selected assignments measurably reduce error rather than emerging incidentally during training.

\subsection{ROCOM}
\Tabref{tab:routing_ROCOM} presents ROCOM results.
The $cboron$, $rhol$, $vl_{0,1,2}$, $mutl$, and $yplus$ are routed to the attention path, while $p$ is routed to the non-attention path.
\input{_tab/Sup_tab_only_switch_results/Path_ROCOM}

\subsection{CUPID}
\Tabref{tab:routing_CUPID} presents CUPID results.
The $qvol\_mas$, $tl$, $vl_{0,1,2}$, and $twall\_rv1\sim5$ are routed to the attention path, while $p$ is routed to the non-attention path.
\input{_tab/Sup_tab_only_switch_results/Path_CUPID}

\subsection{Synthetic}
\Tabref{tab:routing_Synthetic} presents Synthetic results.
All four variables ($\epsilon$, $k$, $\mu_t$, $p$) are routed to the attention path, with no variable assigned to the non-attention path.
\input{_tab/Sup_tab_only_switch_results/Path_Synthetic}

\input{_fig/fig_tex/fig8_bit_comparison}

\subsection{SMR Simulation}
\Tabref{tab:routing_SMR} presents SMR Simulation results.
The $p$, $U$, $V$, and $\omega_z$ are routed to the attention path, while $W$, $\omega_x$, $\omega_y$, and $L2$ are routed to the non-attention path.
\input{_tab/Sup_tab_only_switch_results/Path_SMR}

\subsection{Combustion}
\Tabref{tab:routing_combustion} presents Combustion results.
The reactive scalars ($jet_{mixfrac}$, $jet_{Y_{OH}}$, $jet_{hr}$) are routed to the attention path, while $jet_{vort}$ is routed to the non-attention path.
\input{_tab/Sup_tab_only_switch_results/Path_Combustion}

\subsection{Ionization}
\Tabref{tab:routing_ionization} presents Ionization results.
All four variables ($H_2$, $H^+$, $He$, $PD$) are routed to the attention path, with no variable assigned to the non-attention path.
\input{_tab/Sup_tab_only_switch_results/Path_Ionization}

\input{_tab/Sup_Tab_quantization}

\begin{figure*}[!htbp]
    \centering
    \includegraphics[width=0.99\textwidth]{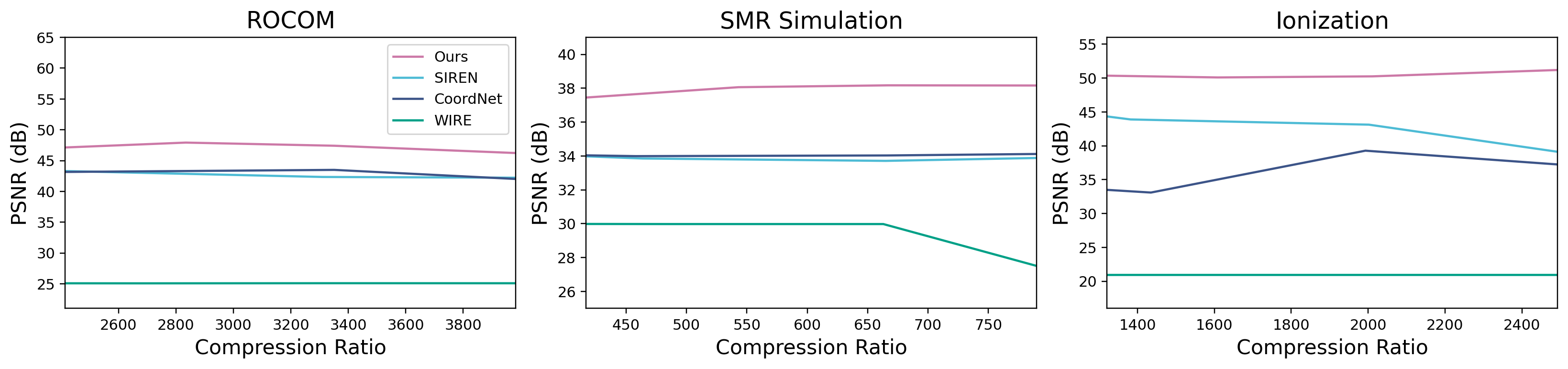}   
    \caption{
    PSNR graph across various CRs for SIREN, CoordNet, WIRE, and Ours on ROCOM, SMR Simulation, and Ionization.
    }
    \label{fig:PSNR_CR}
\end{figure*}

\section{Quantization}
\label{sec:add_quan}

We further analyze the effect of 16-bit quantization.
\Figref{fig:bit_comparison} compares the 32-bit and 16-bit reconstructions of SWAP-INR across all six datasets.
The two are visually indistinguishable, confirming that half-precision conversion preserves fine physical details while roughly halving the model size, directly enabling the compression gains reported throughout the paper.
This resilience reflects a known property of coordinate-based neural representations, which are largely robust to reduced precision.

To verify that SWAP-INR's advantage stems from its representation rather than from quantization alone,~\tabref{tab:16bit_exp} applies the same 16-bit conversion to all competing methods, placing every baseline on equal footing at matched CR.
Note that WIRE is excluded, as its complex-valued Gabor-wavelet activations are incompatible with real-valued 16-bit conversion.
Even under this uniform 16-bit setting, SWAP-INR achieves SOTA performance on the SMR datasets (ROCOM~\cite{cho2019numerical}, CUPID, Synthetic, and SMR Simulation) and remains competitive on the public datasets (Combustion~\cite{hawkes2007scalar} and Ionization~\cite{whalen2008ionization}) at matched CR.
This confirms that the advantage of SWAP-INR is intrinsic to its architectural design rather than to numerical precision, persisting even when every competing baseline benefits equally from the same quantization.

\section{PSNR Results under Various CR}

\Figref{fig:PSNR_CR} illustrates how PSNR changes as the CR increases for SIREN, CoordNet, WIRE, and our proposed SWAP-INR across three datasets.
Across all evaluations, SWAP-INR consistently and significantly outperforms the baselines at every compression level, while WIRE remains the weakest throughout, reflecting the difficulty its wavelet activations face in these scientific fields.
Beyond this overall margin, the trends reveal distinct behaviors per dataset.
On ROCOM~\cite{cho2019numerical}, SWAP-INR is remarkably stable, holding a peak PSNR near 48~dB; although it degrades slightly beyond CR $>$ 3200, it still retains a wide fidelity margin over the baselines.
On SMR Simulation, SWAP-INR is notably robust to increasing compression: rather than the steady decline expected at higher CR, its PSNR stays essentially flat, indicating that our targeted pruning removes redundant parameters with little impact on the retained representation.
On Ionization~\cite{whalen2008ionization}, PSNR even rises slightly beyond CR $\approx$ 2000, showing that SWAP-INR sustains, and can modestly improve, fidelity under aggressive compression.
This stability is the key practical advantage of SWAP-INR: it sustains high-fidelity reconstruction even under aggressive compression, where baseline methods either degrade or plateau at substantially lower quality.
Together, these results establish that SWAP-INR achieves extreme compression while preserving the fidelity of complex physical fields.

%% file: _tab/Tab2_comparison_model.tex
\begin{table}[!ht]
\caption{
Structural comparison of mainstream INR-based and traditional compressors. 
Multi.: native multivariate, Grid-agn.: grid-agnostic, No-Interf: interference-free, Corr.: explicit correlation, Impl.: implementation.
}
\label{tab:methods}
\centering
\resizebox{0.99\linewidth}{!}{
\begin{tabular}{l|ccccc}
\Xhline{2\arrayrulewidth}
Methods & Multi. & Grid-agn. & No-Interf. & Corr. & Impl. \\ \hline \hline
ZFP~\cite{lindstrom2014fixed}         & Yes & No & Yes & No & Official \\
TTHRESH~\cite{ballester2019tthresh}  & Yes & No & Yes & No & Official \\
SZ3~\cite{liang2022sz3}               & Yes & Yes & No & No & Official \\
\hline
SIREN~\cite{sitzmann2020implicit}     & No & Yes & No & No & Official \\
NeurComp~\cite{lu2021compressive}     & No & No  & No & No & Official \\
CoordNet~\cite{han2022coordnet}       & No & Yes & No & No & Official \\
Instant-NGP~\cite{muller2022instant}  & No & Yes & No & No & Official \\
fV-SRN~\cite{weiss2022fast}           & No & Yes & No & No & Official \\
WIRE~\cite{saragadam2023wire}         & No & Yes & No & No & Official \\
AFNet~\cite{zhang2023frequency}       & No & Yes & No & No & Official \\
ECNR~\cite{tang2023ecnr}              & No & No  & No & No & None \\
KD-INR~\cite{han2023kd}               & No & Yes & No & No & None \\
APMGSRN~\cite{wurster2023adaptively}  & No & Yes & No & No & Official \\
RMDSRN~\cite{xiong2024regularized}    & No & No  & No & No & None \\
SL$^2$A-INR~\cite{rezaeian2025sl2a}   & No & Yes & No & No & Official \\
DC-INR~\cite{han2025dcinr}            & No & No  & No & No & None \\
MC-INR~\cite{son2025mc}               & Yes & Yes & Yes & No & Official \\
MoE-INR~\cite{han2026moe}             & No & Yes & No & No & None \\
\hline
\textbf{Ours (SWAP-INR)}              & Yes & Yes & Yes & Yes & --- \\
\hline
\Xhline{2\arrayrulewidth}
\end{tabular}
}
\end{table}

%% file: _tab/Tab4_compressor.tex
\begin{table}[ht]
\centering
\caption{Quantitative comparison of compression results with conventional lossy compressors at matched PSNR levels.
    The $\ddagger$ is grid-dependent methods evaluated by converting the unstructured to structured grids.
    }
\label{tab:compression}

\resizebox{0.8\linewidth}{!}{
\begin{tabular}{c|c|c|cc}
\Xhline{2\arrayrulewidth}
Dataset & Method & PSNR & CR$\uparrow$ & Size(KB)$\downarrow$ \\
\hline
\hline

\multirow{4}{*}{ROCOM}
& SZ3                    & \multirow{4}{*}{47.41} & 14 & 141,163  \\
& ZFP$^\ddagger$         &                        & 11 & 171,308  \\
& TTHRESH$^\ddagger$     &                        & \underline{169} & \underline{11,510}   \\
& \textbf{Ours}          &                        & \textbf{3,351} & \textbf{580}      \\
\hline

\multirow{4}{*}{CUPID}
& SZ3                    & \multirow{4}{*}{50.29} & 26 & 63,001   \\
& ZFP$^\ddagger$         &                        & 15 & 108,906  \\
& TTHRESH$^\ddagger$     &                        & \underline{212} & \underline{7,783}    \\
& \textbf{Ours}          &                        & \textbf{2,634} & \textbf{626}      \\
\hline

\multirow{4}{*}{Synthetic}
& SZ3                    & \multirow{4}{*}{75.00} & 39 & 63,298   \\
& ZFP$^\ddagger$         &                        & 20 & 124,080  \\
& TTHRESH$^\ddagger$     &                        & \textbf{18,799} & \textbf{131}      \\
& \textbf{Ours}          &                        & \underline{7,868} & \underline{313}      \\
\hline

\multirow{4}{*}{\begin{tabular}{@{}c@{}}SMR \\ Simulation\end{tabular}}
& SZ3                    & \multirow{4}{*}{38.17} & \underline{36} & \underline{9,114}    \\
& ZFP                    &                        & 13 & 24,584   \\
& TTHRESH                &                        & 18 & 18,701   \\
& \textbf{Ours}          &                        & \textbf{667} & \textbf{491}      \\
\hline

\multirow{4}{*}{Combustion}
& SZ3                    & \multirow{4}{*}{31.17} & 44 & 5,799    \\
& ZFP                    &                        & 34 & 7,420    \\
& TTHRESH                &                        & \underline{72} & \underline{3,492}    \\
& \textbf{Ours}          &                        & \textbf{997} & \textbf{254}      \\
\hline

\multirow{4}{*}{Ionization}
& SZ3                    & \multirow{4}{*}{50.25} & 102 & 4,410    \\
& ZFP                    &                        & 9 & 51,996   \\
& TTHRESH                &                        & \underline{258} & \underline{1,748}    \\
& \textbf{Ours}          &                        & \textbf{2,011} & \textbf{224}      \\

\Xhline{2\arrayrulewidth}

\end{tabular}
}

\end{table}

%% file: _fig/fig_tex/fig7_compressor_comparison.tex
\begin{figure*}[t]
    \centering
    \includegraphics[width=0.99\linewidth]{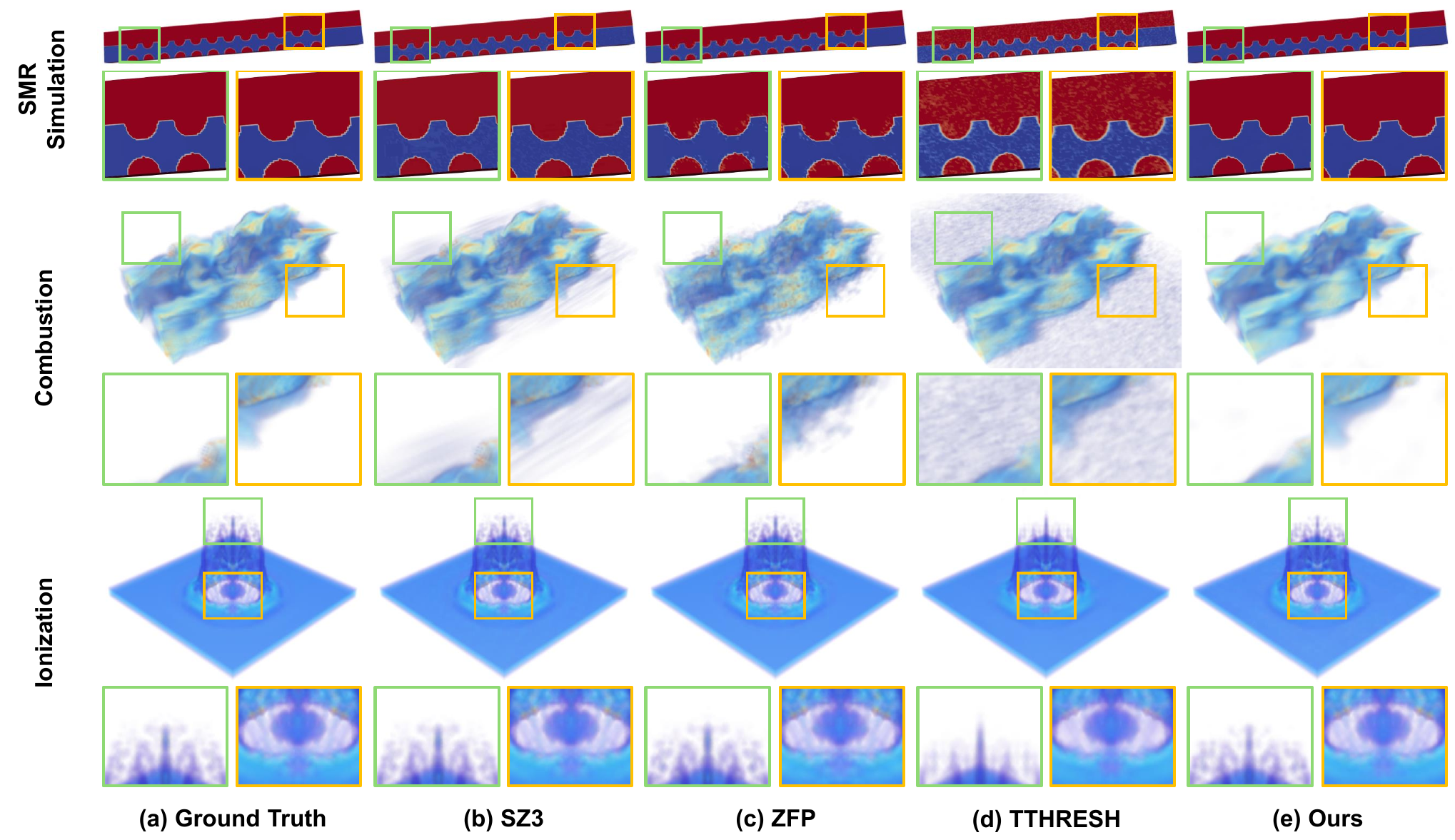}
    \caption{
    Visualization of volume at matched PSNR.
    The volume is reconstructed for a specific snapshot (time and variable).
    The green and orange boxes, along with their zoomed-in highlight regions, where SWAP-INR excels at compression quality.
    }
    \label{fig:compressor_comparison}
\end{figure*}

%% file: _fig/fig_tex/fig4_gradient.tex
\begin{figure}[!ht]
    \centering
    \includegraphics[width=0.97\linewidth]{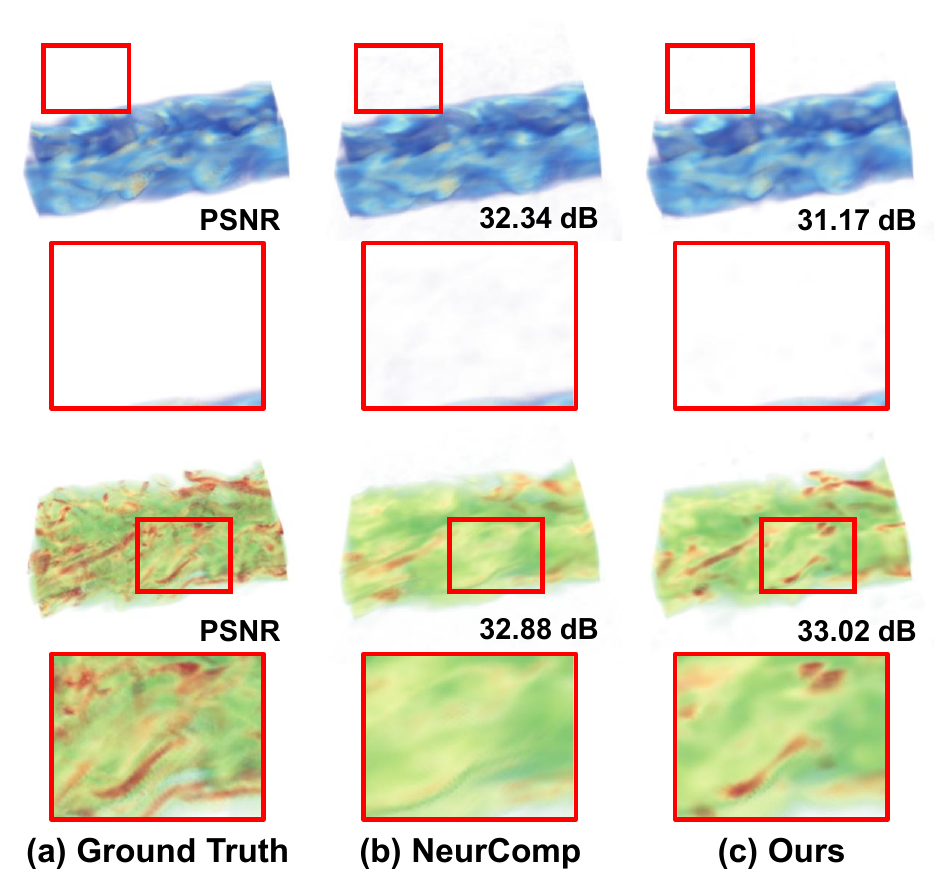}
    \caption{
    Comparison with NeurComp on the Combustion dataset.
    (Top) The $jet_{hr}$ variable: Reconstructions via the attention path
    While our method exhibits slight blurring and a lower PSNR due to the weighted averaging effect, it significantly suppresses structural artifacts.
    (Bottom) The $jet_{vort}$ variable: Reconstructions via the non-attention path.
    Here, our method outperforms NeurComp by 0.14 dB, better capturing high-frequency details.
    Red boxes indicate zoomed-in regions highlighting these improvements.
    }
    \label{fig:vs_neurcomp}
\end{figure}

%% file: _fig/fig_tex/fig_ionization.tex
\begin{figure}[htbp]
    \centering
    \includegraphics[width=0.97\linewidth]{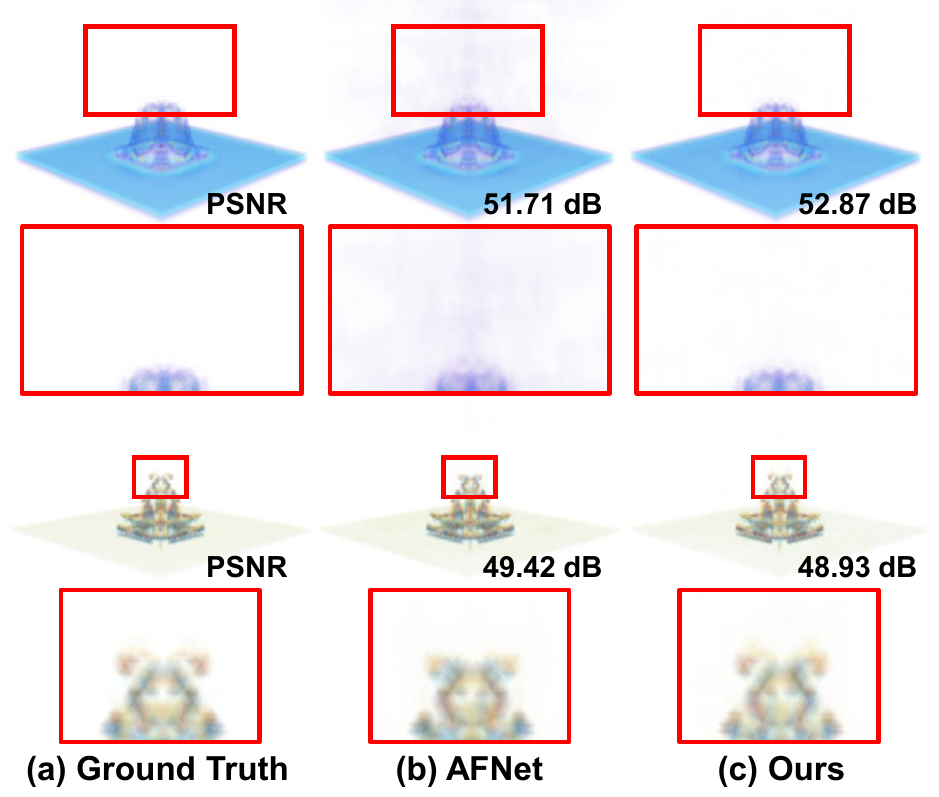}
    \caption{
    Comparison with AFNet on the Ionization dataset.
    (Top) The $PD$ variable: AFNet introduces noticeable noise across the volume, whereas the low-pass nature of attention suppresses this noise, yielding clean, artifact-free reconstructions.
    (Bottom) The $He$ variable, which benefits from cross-variable correlation, although attention induces slight blurring, ours reconstructs the underlying structure more faithfully than AFNet.
    Red boxes indicate zoomed-in regions highlighting these improvements.
    }
    \label{fig:vs_afnet}
\end{figure}

%% file: _fig/fig_tex/fig5_hotcorner.tex
\begin{figure}[!ht]
    \centering
    \includegraphics[width=0.95\linewidth]{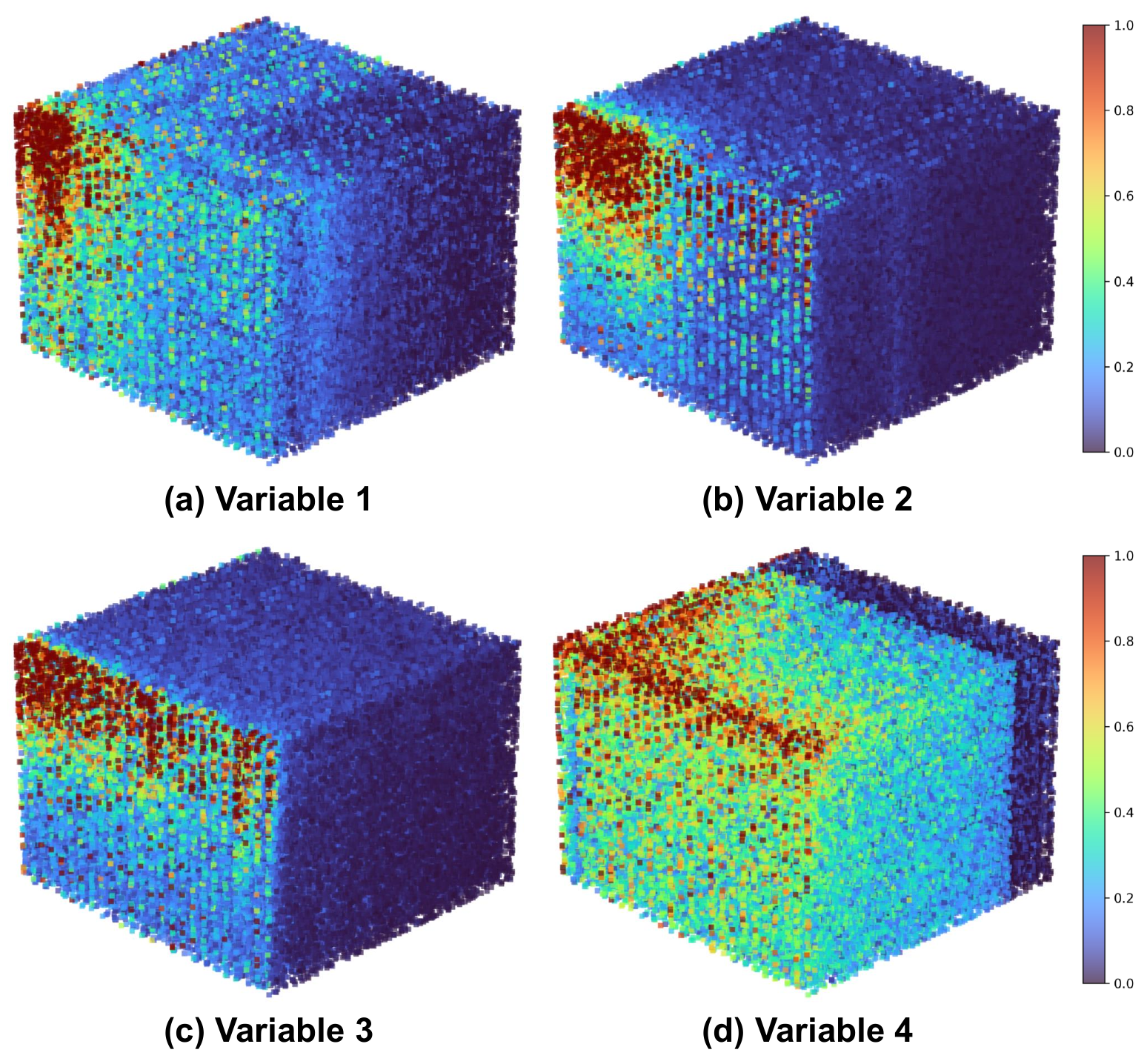}
    \caption{
    Visualization of the HOSVD core tensor for the Synthetic dataset, demonstrating the hot corner phenomenon.
    %
    Across all variables, the colormap represents the normalized structural weight of the tensor entries.
    blue: discardable background noise, red: core structural components.
    }
    \label{fig:hotcorner}
\end{figure}

%% file: _tab/Sup_Tab_time.tex
\begin{table*}[htbp]
\centering
\caption{Comparison of training and inference times across datasets.
$t$ (sec) and $t$ (hr) denote the time per epoch and total training time, respectively, and Inf. (sec) denotes the full-volume inference time.
}
\label{tab:time}
\resizebox{0.99\linewidth}{!}{
\begin{tabular}{c|c|ccc|c|c|ccc|c|c|ccc}
\Xhline{2\arrayrulewidth}
Dataset & Method & $t$ (sec) & $t$ (hr) & Inf. (sec) & Dataset & Method & $t$ (sec) & $t$ (hr) & Inf. (sec) & Dataset & Method & $t$ (sec) & $t$ (hr) & Inf. (sec) \\
\hline
\hline
\multirow{11}{*}{ROCOM}
& SIREN        & 216 & 36.2  & 177 & \multirow{11}{*}{CUPID}
& SIREN        & 140 & 4.7   & 136 & \multirow{11}{*}{Synthetic}
& SIREN        & 600 & 20.3  & 483 \\
& NeurComp     & 223 & 13.8  & 176 &
& NeurComp     & 138 & 7.0   & 109 &
& NeurComp     & 533 & 26.5  & 466 \\
& CoordNet     & 227 & 45.4  & 182 &
& CoordNet     & 134 & 15.3  & 150 &
& CoordNet     & 686 & 31.5  & 494 \\
& Instant-NGP  & 199 & 137.6 & 205 &
& Instant-NGP  & 133 & 85.1  & 124 &
& Instant-NGP  & 524 & 185.5 & 494 \\
& fV-SRN       & 230 & 132.7 & 200 &
& fV-SRN       & 240 & 109.7 & 210 &
& fV-SRN       & 600 & 338.3 & 656 \\
& WIRE         & 246 & 8.2   & 198 &
& WIRE         & 132 & 6.7   & 142 &
& WIRE         & 598 & 19.9  & 542 \\
& APMGSRN      & 210 & 128.7 & 197 &
& APMGSRN      & 180 & 113.3 & 209 &
& APMGSRN      & 630 & 346.1 & 654 \\
& AFNet        & 226 & 125.6 & 178 &
& AFNet        & 126 & 58.0  & 137 &
& AFNet        & 593 & 137.8 & 481 \\
& SL$^2$A-INR  & 225 & 125.0 & 179 &
& SL$^2$A-INR  & 134 & 74.4  & 145 &
& SL$^2$A-INR  & 547 & 158.3 & 464 \\
& MC-INR       & 618 & 53.2  & 202 &
& MC-INR       & 301 & 38.5  & 282 &
& MC-INR       & 726 & 170.6 & 641 \\
& \textbf{Ours}& 210 & 33.4  & 193 &
& \textbf{Ours}& 136 & 9.5   & 132 &
& \textbf{Ours}& 517 & 34.0  & 483 \\
\hline
\hline
\multirow{11}{*}{\begin{tabular}{@{}c@{}}SMR \\ Simulation\end{tabular}}
& SIREN        & 39  & 2.3   & 37  & \multirow{11}{*}{Combustion}
& SIREN        & 59  & 11.8  & 51  & \multirow{11}{*}{Ionization}
& SIREN        & 92  & 5.6   & 82  \\
& NeurComp     & 96  & 19.8  & 40  &
& NeurComp     & 61  & 11.8  & 49  &
& NeurComp     & 110 & 6.0   & 87  \\
& CoordNet     & 41  & 3.4   & 32  &
& CoordNet     & 63  & 12.9  & 48  &
& CoordNet     & 101 & 5.2   & 81  \\
& Instant-NGP  & 35  & 19.4  & 36  &
& Instant-NGP  & 61  & 33.2  & 53  &
& Instant-NGP  & 95  & 58.3  & 93  \\
& fV-SRN       & 60  & 28.9  & 61  &
& fV-SRN       & 75  & 44.7  & 144 &
& fV-SRN       & 150 & 85.1  & 263 \\
& WIRE         & 39  & 5.0   & 33  &
& WIRE         & 62  & 6.1   & 47  &
& WIRE         & 103 & 5.9   & 90  \\
& APMGSRN      & 50  & 28.0  & 64  &
& APMGSRN      & 75  & 44.6  & 130 &
& APMGSRN      & 150 & 93.1  & 265 \\
& AFNet        & 36  & 28.8  & 32  &
& AFNet        & 57  & 31.6  & 52  &
& AFNet        & 106 & 54.7  & 94  \\
& SL$^2$A-INR  & 39  & 12.2  & 32  &
& SL$^2$A-INR  & 63  & 35.0  & 46  &
& SL$^2$A-INR  & 107 & 48.2  & 83  \\
& MC-INR       & 58  & 4.6   & 59  &
& MC-INR       & 154 & 19.4  & 44  &
& MC-INR       & 142 & 11.2  & 79  \\
& \textbf{Ours}& 42  & 4.4   & 39  &
& \textbf{Ours}& 56  & 49.3  & 56  &
& \textbf{Ours}& 98  & 6.8   & 88  \\
\Xhline{2\arrayrulewidth}
\end{tabular}
}
\end{table*}

%% file: _tab/Sup_tab_only_switch_results/Path_ROCOM.tex
\begin{table}[htbp]
    \centering
    \caption{Path switching results for the ROCOM dataset~\cite{cho2019numerical}.}
    \label{tab:routing_ROCOM}
    \resizebox{0.99\columnwidth}{!}{
    \begin{tabular}{lccc}
    \toprule
    \textbf{Variables} & \textbf{$\text{Path}_{attn}$} & \textbf{$\text{Path}_{non-attn}$} & PSNR$\uparrow$ \\
    \midrule
    $cboron$ (Concentration of Boron)       & \checkmark &            & 67.12 \\
    $mutl$ (Turbulent viscosity)            & \checkmark &            & 32.64 \\
    $p$ (Pressure)                          &            & \checkmark & 53.35 \\
    $rhol$ (Density of liquid)              & \checkmark &            & 53.40 \\
    $vl_{0}$ (Liquid velocity vector)       & \checkmark &            & 43.14 \\
    $vl_{1}$ (Liquid velocity vector)       & \checkmark &            & 46.65 \\
    $vl_{2}$ (Liquid velocity vector)       & \checkmark &            & 45.34 \\
    $yplus$ (Non-dimensional wall distance) & \checkmark &            & 37.67 \\
    \bottomrule
    \end{tabular}
    }
\end{table}

%% file: _tab/Sup_tab_only_switch_results/Path_CUPID.tex
\begin{table}[htbp]
    \centering
    \caption{Path switching results for the CUPID dataset.}
    \label{tab:routing_CUPID}
    \resizebox{0.99\columnwidth}{!}{
    \begin{tabular}{lccc}
    \toprule
    \textbf{Variables} & \textbf{$\text{Path}_{attn}$} & \textbf{$\text{Path}_{non-attn}$} & PSNR$\uparrow$ \\
    \midrule
    $p$ (Pressure)                   &            & \checkmark & 50.37 \\
    $qvol\_mas$ (Axial core power)   & \checkmark &            & 45.39 \\
    $tl$ (Temperature of liquid)     & \checkmark &            & 45.28 \\
    $twall\_rv1$ (User-defined)      & \checkmark &            & 52.76 \\
    $twall\_rv2$ (User-defined)      & \checkmark &            & 45.74 \\
    $twall\_rv3$ (User-defined)      & \checkmark &            & 60.48 \\
    $twall\_rv4$ (User-defined)      & \checkmark &            & 52.03 \\
    $twall\_rv5$ (User-defined)      & \checkmark &            & 50.19 \\
    $vl_0$ (Liquid velocity vector)  & \checkmark &            & 52.95 \\
    $vl_1$ (Liquid velocity vector)  & \checkmark &            & 54.00 \\
    $vl_2$ (Liquid velocity vector)  & \checkmark &            & 43.96 \\
    \bottomrule
    \end{tabular}
    }
\end{table}

%% file: _tab/Sup_tab_only_switch_results/Path_Synthetic.tex
\begin{table}[htbp]
    \centering
    \caption{Path switching results for the Synthetic dataset.}
    \label{tab:routing_Synthetic}
    \resizebox{0.99\columnwidth}{!}{%
    \begin{tabular}{lccc}
    \toprule
    \textbf{Variables} & \textbf{$\text{Path}_{attn}$} & \textbf{$\text{Path}_{non-attn}$} & PSNR$\uparrow$ \\
    \midrule
    $\epsilon$ (Dissipation rate of turbulence) & \checkmark &  & 79.58 \\
    $k$ (Turbulent kinetic energy)              & \checkmark &  & 75.10 \\
    $\mu_t$ (Turbulent viscosity)               & \checkmark &  & 68.70 \\
    $p$ (Pressure)                              & \checkmark &  & 76.60 \\
    \bottomrule
    \end{tabular}
    }
\end{table}

%% file: _fig/fig_tex/fig8_bit_comparison.tex
\begin{figure*}[htbp]
    \centering
    \includegraphics[width=0.94\linewidth]{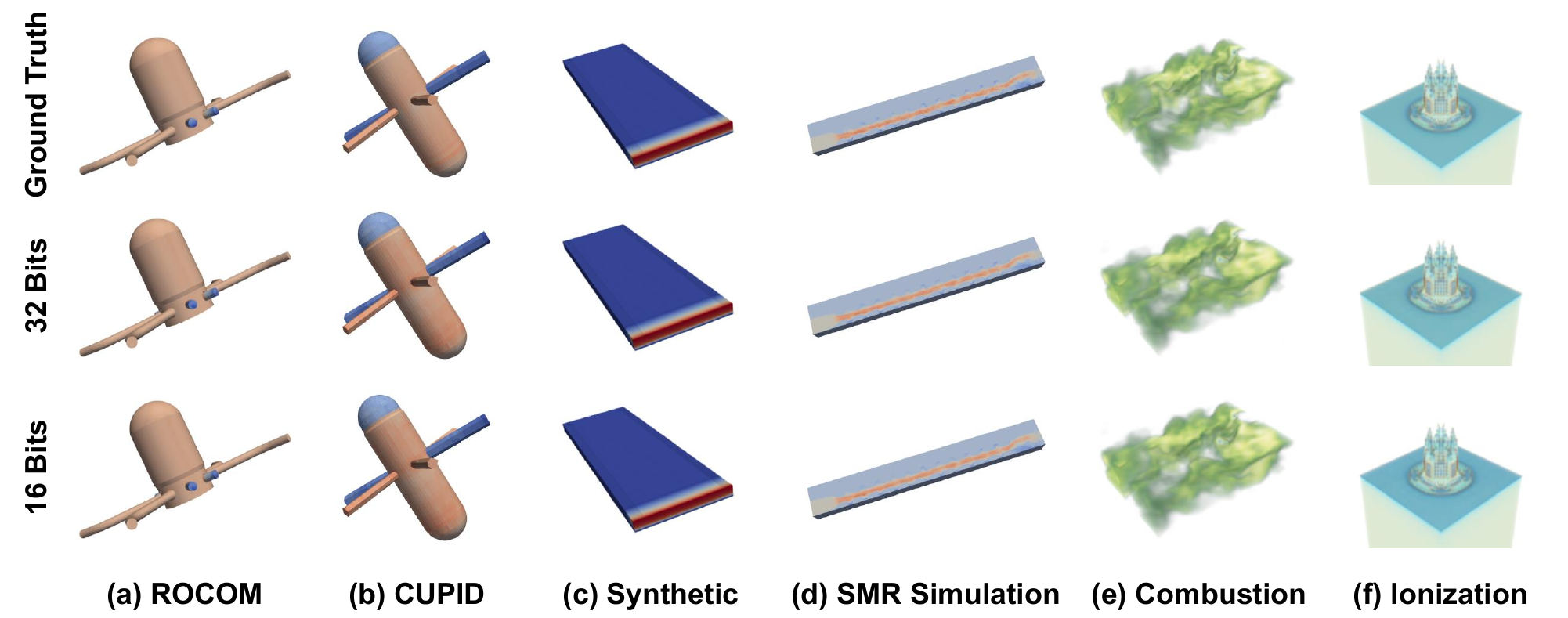}
    \caption{
    Visual comparison of ground truth data (top row) with model reconstructions using 32-bit (middle row) and quantized 16-bit weights (bottom row).
    The volume is rendered for a specific snapshot (time and variable).
    }
    \label{fig:bit_comparison}
\end{figure*}

%% file: _tab/Sup_tab_only_switch_results/Path_SMR.tex
\begin{table}[htbp]
    \centering
    \caption{Path switching results for the SMR Simulation dataset.}
    \label{tab:routing_SMR}

    \resizebox{0.99\columnwidth}{!}{
    \begin{tabular}{lccc}
    \toprule
    \textbf{Variables} & \textbf{$\text{Path}_{attn}$} & \textbf{$\text{Path}_{non-attn}$} & PSNR$\uparrow$ \\
    \midrule
    $L2$ (Vortex identifier)   &            & \checkmark & 48.36 \\
    $p$ (Pressure)                                  & \checkmark &            & 47.90 \\
    $U$ (Velocity $x$ components)                   & \checkmark &            & 32.62 \\
    $V$ (Velocity $y$ components)                   & \checkmark &            & 31.11 \\
    $W$ (Velocity $z$ component)                    &            & \checkmark & 32.22 \\
    $\omega_x$ (Vorticity $x$ components)           &            & \checkmark & 37.01 \\
    $\omega_y$ (Vorticity $y$ components)           &            & \checkmark & 37.45 \\
    $\omega_z$ (Vorticity $z$ component)            & \checkmark &            & 38.68 \\
    \bottomrule
    \end{tabular}
    }
\end{table}

%% file: _tab/Sup_tab_only_switch_results/Path_Combustion.tex
\begin{table}[htbp]
    \centering
    \caption{Path switching results for the Combustion dataset~\cite{hawkes2007scalar}.}
    \label{tab:routing_combustion}
    \resizebox{0.99\columnwidth}{!}{
    \begin{tabular}{lccc}
    \toprule
    \textbf{Variables} & \textbf{$\text{Path}_{attn}$} & \textbf{$\text{Path}_{non-attn}$} & PSNR$\uparrow$ \\
    \midrule
    $jet_{hr}$ (Heat release rate)      & \checkmark &            & 30.80 \\
    $jet_{mixfrac}$ (Mixture fraction)  & \checkmark &            & 30.35 \\
    $jet_{vort}$ (Vorticity)            &            & \checkmark & 32.80 \\
    $jet_{Y_{OH}}$ (OH mass fraction)   & \checkmark &            & 30.74 \\
    \bottomrule
    \end{tabular}
    }
\end{table}

%% file: _tab/Sup_tab_only_switch_results/Path_Ionization.tex
\begin{table}[htbp]
    \centering
    \caption{Path switching results for the Ionization dataset~\cite{whalen2008ionization}.}
    \label{tab:routing_ionization}
    \resizebox{0.99\columnwidth}{!}{
    \begin{tabular}{lccc}
    \toprule
    \textbf{Variables} & \textbf{$\text{Path}_{attn}$} & \textbf{$\text{Path}_{non-attn}$} & PSNR$\uparrow$ \\
    \midrule
    $H_2$ (Molecular hydrogen)       & \checkmark &  & 52.97 \\
    $H^+$ (Ionized hydrogen)         & \checkmark &  & 48.21 \\
    $He$ (Neutral helium)            & \checkmark &  & 46.80 \\
    $PD$ (Photon/Particle density)   & \checkmark &  & 53.02 \\
    \bottomrule
    \end{tabular}
    }
\end{table}

%% file: _tab/Sup_Tab_quantization.tex
\begin{table*}[htbp]
\centering
\caption{
Quantitative comparison with all methods under 16-bit quantization.
}
\label{tab:16bit_exp}
\resizebox{0.99\linewidth}{!}{
\begin{tabular}{c|c|ccccc|c|c|ccccc}
\Xhline{2\arrayrulewidth}
Dataset & Method & PSNR$\uparrow$ & NRMSE$\downarrow$ & $R^2\uparrow$ & CR$\uparrow$ & Size(KB)$\downarrow$
& Dataset & Method & PSNR$\uparrow$ & NRMSE$\downarrow$ & $R^2\uparrow$ & CR$\uparrow$ & Size(KB)$\downarrow$ \\
\hline
\hline

\multirow{11}{*}{ROCOM}
& SIREN              & 42.84 & 0.0105 & 0.9417  & 3,345 & 581             
& \multirow{11}{*}{CUPID}
& SIREN              & 23.62 & 0.0098 & 0.6511  & 2,622 & 629   \\        
& NeurComp           & 42.59 & 0.0105 & 0.9399  & 3,328 & 584             
& & NeurComp         & 23.15 & 0.1163 & 0.6457  & 2,630 & 627  \\         
& CoordNet           & 42.88 & 0.0108 & 0.9382  & 3,323 & 585             
& & CoordNet         & 22.19 & 0.0164 & 0.6110  & 2,626 & 628 \\          
& Instant-NGP        & 32.49 & 0.0266 & 0.5243  & 3,351 & 580             
& & Instant-NGP      & 19.98 & 0.1759 & 0.4335  & 2,561 & 644 \\          
& fV-SRN$^\ast$      & 41.45 & 0.0112 & 0.9145  & 3,351 & 580             
& & fV-SRN$^\ast$    & 21.52 & 0.1616 & 0.4151  & 2,618 & 630 \\          
& WIRE               & --    & --     & --      & --    & --              
& & WIRE             & --    & --     & --      & --    & --  \\          
& AFNet              & \underline{44.98} & \underline{0.0087} & \underline{0.9545}  & 3,351 & 580             
& & AFNet            & \underline{49.95} & \underline{0.0036} & \underline{0.9779}  & 2,622 & 629 \\          
& APMGSRN$^\ast$     & 39.27 & 0.0161 & 0.8844  & 3,351 & 580             
& & APMGSRN$^\ast$   & 20.20 & 0.1723 & 0.2197  & 2,618 & 630 \\          
& SL$^2$A-INR        & 41.33 & 0.0123 & 0.9254  & 3,328 & 584             
& & SL$^2$A-INR      & 22.45 & 0.1186 & 0.5282  & 2,622 & 629 \\          
& MC-INR$^\ast$      & 39.78 & 0.0168 & 0.8360  & 3,328 & 584             
& & MC-INR$^\ast$    & 46.14 & 0.0078 & 0.9647  & 2,626 & 628 \\          
& \textbf{Ours}  & \textbf{47.41} & \textbf{0.0073} & \textbf{0.9782} & 3,351  & 580        
& & \textbf{Ours}& \textbf{50.29} & \textbf{0.0035} & \textbf{0.9904} & 2,634  & 626  \\    

\hline
\hline

\multirow{11}{*}{Synthetic}
& SIREN              & 65.11 & 0.0007 & 0.8558  & 7,818 & 315  
& \multirow{11}{*}{\begin{tabular}{@{}c@{}}SMR \\ Simulation\end{tabular}}
& SIREN              & 34.21 & 0.0248 & 0.3192  & 667   & 491 \\          
& NeurComp           & 68.06 & 0.0004 & 0.7237  & 7,769 & 317             
& & NeurComp         & 33.36 & 0.0287 & 0.2119  & 666   & 492 \\          
& CoordNet           & 63.73 & 0.0011 & \underline{0.9731}  & 7,868 & 313             
& & CoordNet         & 34.14 & 0.0259 & 0.3427  & 658   & 498 \\          
& Instant-NGP        & 35.22 & 0.0772 & 0.0514  & 7,601 & 324             
& & Instant-NGP      & 30.27 & 0.0811 & 0.0565  & 635   & 516  \\         
& fV-SRN$^\ast$      & 66.68 & 0.0005 & 0.8526  & 7,463 & 330             
& & fV-SRN$^\ast$    & 30.61 & 0.0359 & -2.7409 & 655   & 500  \\         
& WIRE               & --    & --     & --      & --    & --              
& & WIRE             & --    & --     & --      & --    & --  \\          
& AFNet              & \underline{71.69} & \underline{0.0003} & 0.9379  & 7,793 & 316             
& & AFNet            & \underline{36.24} & \underline{0.0178} & \underline{0.3907}  & 667   & 491 \\          
& APMGSRN$^\ast$     & 64.46 & 0.0007 & 0.8124  & 7,463 & 330             
& & APMGSRN$^\ast$   & 30.90 & 0.0512 & -0.3559 & 655   & 500  \\         
& SL$^2$A-INR        & 41.41 & 0.0086 & -121.0300 & 7,843 & 314             
& & SL$^2$A-INR      & 31.58 & 0.0354 & -1.2366 & 665   & 493 \\          
& MC-INR             & 65.03 & 0.0007 & 0.9507  & 7,696 & 320             
& & MC-INR           & 34.03 & 0.0250 & 0.2843  & 650   & 504  \\         
& \textbf{Ours}  & \textbf{75.00} & \textbf{0.0002} & \textbf{0.9736} & 7,868 & 313 
& & \textbf{Ours}& \textbf{38.17} & \textbf{0.0153} & \textbf{0.4494} & 667 & 491 \\ 

\hline
\hline

\multirow{11}{*}{Combustion}
& SIREN               & 31.06 & 0.0281 & 0.9057  & 989 & 256              
& \multirow{11}{*}{Ionization}
& SIREN               & 38.34 & 0.0162 & 0.9402  & 2,011 & 224   \\       
& NeurComp            & \textbf{32.86} & \textbf{0.0227} & \textbf{0.9197}  & 997   & 254            
& & NeurComp          & 39.87 & 0.0129 & 0.9526  & 2,011 & 224   \\       
& CoordNet            & 30.74 & 0.0292 & 0.9030  & 985   & 257            
& & CoordNet          & 32.75 & 0.0321 & 0.8128  & 1,993 & 226  \\        
& Instant-NGP         & 21.03 & 0.1009 & 0.4953  & 977   & 259            
& & Instant-NGP       & 22.47 & 0.1767 & 0.2545  & 1,984 & 227 \\         
& fV-SRN$^\ast$       & 24.91 & 0.0607 & 0.7590  & 920   & 275            
& & fV-SRN$^\ast$     & 26.48 & 0.0657 & 0.2457  & 1,802 & 250 \\         
& WIRE                & --    & --     & --      & --    & --             
& & WIRE              & --    & --     & --      & --    & --  \\         
& AFNet               & 31.03 & 0.0282 & 0.9022  & 993   & 255            
& & AFNet             & \underline{48.64} & \underline{0.0044} & \textbf{0.9916}  & 1,984 & 227 \\         
& APMGSRN$^\ast$      & 22.67 & 0.0808 & 0.6272  & 920   & 275            
& & APMGSRN$^\ast$    & 19.77 & 0.1306 & -3.8897 & 1,802 & 250 \\         
& SL$^2$A-INR         & 24.21 & 0.0669 & 0.7251  & 997   & 254            
& & SL$^2$A-INR       & 33.50 & 0.0277 & 0.8109  & 1,984 & 227 \\         
& MC-INR              & 17.46 & 0.1592 & -0.1311 & 989  & 256             
& & MC-INR            & 21.44 & 0.1443 & -0.4746 & 1,942 & 232  \\        
& \textbf{Ours}   & \underline{31.17} & \underline{0.0278} & \underline{0.9116} & 997 & 254   
& & \textbf{Ours} & \textbf{50.25} & \textbf{0.0032} & \underline{0.9905}  & 2,011  & 224  \\        

\Xhline{2\arrayrulewidth}
\end{tabular}
}
\end{table*}